\documentclass{article}
\PassOptionsToPackage{numbers,compress}{natbib}
\PassOptionsToPackage{table}{xcolor}
\usepackage[preprint]{neurips_2026}

\usepackage{charter}
\usepackage{xcolor}
\definecolor{rnSkyCyan}{HTML}{5FB3DC}
\definecolor{rnBlue}{HTML}{3B82F6}
\definecolor{rnBlueDark}{HTML}{1E40AF}
\definecolor{rnBlueLight}{HTML}{E8F1FE}
\definecolor{rnGold}{HTML}{B8860B}
\definecolor{rnGoldLight}{HTML}{F8EFD3}
\definecolor{rnSlate}{HTML}{6B7280}
\definecolor{rnDark}{HTML}{1F2328}

\usepackage{hyperref}
\hypersetup{
  colorlinks=true,
  linkcolor=rnBlueDark,
  citecolor=rnBlueDark,
  urlcolor=rnBlueDark,
  pdfborder={0 0 0}
}
\usepackage{tcolorbox}
\tcbuselibrary{skins,breakable}
\usepackage{graphicx}
\usepackage{booktabs}

\makeatletter
\newlength{\rnHeroSpacing}
\newcommand{\rn@gap}{\par\vspace{\rnHeroSpacing}}
\newcommand{\rn@smallgap}{\par\vspace{0.67\rnHeroSpacing}}

\def\rn@logo{}
\def\rn@logoWidth{0.55\linewidth}
\def\rn@code{}
\def\rn@website{}
\def\rn@heroabstract{}
\def\rn@affiliations{}
\newcommand{\herologo}[1]{\def\rn@logo{#1}}
\newcommand{\herologoWidth}[1]{\def\rn@logoWidth{#1}}
\newcommand{\code}[1]{\def\rn@code{#1}}
\newcommand{\website}[1]{\def\rn@website{#1}}
\newcommand{\heroabstract}[1]{\def\rn@heroabstract{#1}}
\newcommand{\affiliations}[1]{\def\rn@affiliations{#1}}

\newcommand*{\rn@authorsep}{\hspace{0.45em}\textperiodcentered\hspace{0.45em}}

\renewcommand{\@maketitle}{%
  \vbox{%
    \hsize\textwidth
    \linewidth\hsize
    \vskip 0.02in
    \begin{tcolorbox}[
      enhanced,
      colback=rnSkyCyan, opacityback=0.10,
      colframe=rnSkyCyan, opacityframe=0.10,
      boxrule=0pt, arc=6pt,
      left=20pt, right=20pt, top=12pt, bottom=10pt,
      before skip=0pt, after skip=0pt,
      width=\linewidth,
    ]
      \ifx\rn@logo\@empty\else
        \begin{center}
          \includegraphics[width=\rn@logoWidth]{\rn@logo}%
        \end{center}
        \rn@smallgap
      \fi
      \ifx\@title\@empty\else
        \begin{center}
          {\LARGE\bfseries\color{rnDark}%
           \hyphenpenalty=10000\exhyphenpenalty=10000%
           \@title\par}%
        \end{center}
        \rn@smallgap
      \fi
      \ifx\@author\@empty\else
        \begin{center}
          \begingroup
            \def\and{\rn@authorsep}%
            \def\And{\rn@authorsep}%
            \def\AND{\rn@authorsep}%
            {\color{rnDark}\@author\par}%
          \endgroup
        \end{center}
        \rn@smallgap
      \fi
      \ifx\rn@affiliations\@empty\else
        \begin{center}
          {\small\color{rnDark}\rn@affiliations\par}%
        \end{center}
        \rn@smallgap
      \fi
      {\color{rnDark!25}\rule{\linewidth}{0.4pt}}
      \rn@gap
      \ifx\rn@heroabstract\@empty\else
        {\small\color{rnDark}\rn@heroabstract\par}%
        \rn@gap
      \fi
      {\color{rnDark!25}\rule{\linewidth}{0.4pt}}
      \rn@gap
      \def\rn@hasmeta{0}%
      \ifx\rn@code\@empty\else\def\rn@hasmeta{1}\fi
      \ifx\rn@website\@empty\else\def\rn@hasmeta{1}\fi
      \ifx\@date\@empty\else\def\rn@hasmeta{1}\fi
      \if\rn@hasmeta1
        \begingroup
          \setlength{\parskip}{0pt}%
          \setlength{\baselineskip}{14pt}%
          \raggedright\small\color{rnDark}%
          \def\rn@first{1}%
          \ifx\rn@code\@empty\else
            \noindent\textbf{Code:}~\href{\rn@code}{\rn@code}%
            \def\rn@first{0}%
          \fi
          \ifx\rn@website\@empty\else
            \if\rn@first0\\\fi
            \textbf{Website:}~\href{\rn@website}{\rn@website}%
            \def\rn@first{0}%
          \fi
          \ifx\@date\@empty\else
            \if\rn@first0\\\fi
            \textbf{Date:}~\@date%
          \fi
          \par
        \endgroup
      \fi
    \end{tcolorbox}
    \vskip 0.12in \@minus 0.05in
  }
}

\newtcolorbox{rnnote}[1][]{
  enhanced, breakable,
  colback=rnBlueLight, colframe=rnBlueLight,
  fonttitle=\sffamily\bfseries,
  coltitle=rnDark,
  boxrule=0pt, arc=2pt,
  left=10pt, right=10pt, top=8pt, bottom=8pt,
  before skip=10pt, after skip=10pt,
  #1
}

\newtcolorbox{rngold}[1][]{
  enhanced, breakable,
  colback=rnGoldLight, colframe=rnGoldLight,
  fonttitle=\sffamily\bfseries,
  coltitle=rnDark,
  boxrule=0pt, arc=2pt,
  left=10pt, right=10pt, top=8pt, bottom=8pt,
  before skip=10pt, after skip=10pt,
  #1
}
\makeatother

\usepackage[utf8]{inputenc} %
\usepackage[T1]{fontenc}    %
\usepackage{url}            %
\usepackage{amsfonts}       %
\usepackage{nicefrac}       %
\usepackage{microtype}      %
\usepackage{enumitem}       %

\usepackage{amsmath,amsfonts,bm}

\def\eqref#1{equation~\ref{#1}}

\def\1{\bm{1}}

\DeclareMathAlphabet{\mathsfit}{\encodingdefault}{\sfdefault}{m}{sl}
\SetMathAlphabet{\mathsfit}{bold}{\encodingdefault}{\sfdefault}{bx}{n}

\usepackage{hyperref}
\usepackage{url}
\usepackage{listings}
\usepackage{multirow}
\usepackage[capitalize,noabbrev]{cleveref}
\usepackage{booktabs} %
\usepackage{amssymb}  %
\usepackage{graphicx}
\usepackage{amsmath}
\usepackage{xspace}
\usepackage{makecell}
\usepackage{textcomp}

\newcommand{\ojSpec}{\emph{spec}\xspace}
\newcommand{\sysname}{\textsc{OpenJarvis}\xspace}
\newcommand{\methodname}{LLM-guided spec search\xspace}
\newcommand{\ojSpecs}{\emph{specs}\xspace}

\newcommand{\claudeOpus}{\texttt{Claude~Opus~4.6}\xspace}

\usepackage{tcolorbox}
\tcbuselibrary{skins, breakable}
\usepackage{algorithm}
\usepackage{algpseudocode}
\usepackage{xcolor}

\usepackage{booktabs}
\usepackage{siunitx}

\usepackage{wrapfig}
\usepackage{hyperref}

\definecolor{darkgreen}{RGB}{0,100,0}
\definecolor{darkred}{RGB}{139,0,0}

\usepackage[textsize=tiny]{todonotes}

\definecolor{clrgp}{rgb}{.9,0,.9}
\definecolor{red}{rgb}{.8,0,0}
\definecolor{blue}{rgb}{0,0, 0.8}
\definecolor{gray}{rgb}{0.41, 0.41, 0.41}
\definecolor{forestgreen}{rgb}{0.13, 0.55, 0.13}
\definecolor{subtle}{RGB}{152,78,163}

\lstdefinelanguage{TOML}{
  morekeywords={true, false},
  keywordstyle=\color{blue!70!black}\bfseries,
  morestring=[b]",
  stringstyle=\color{purple!60!black},
  morecomment=[l]\#,
  commentstyle=\color{gray}\itshape,
  sensitive=true,
  alsoletter={-_.},
}

\title{Personal AI, On Personal Devices}
\herologo{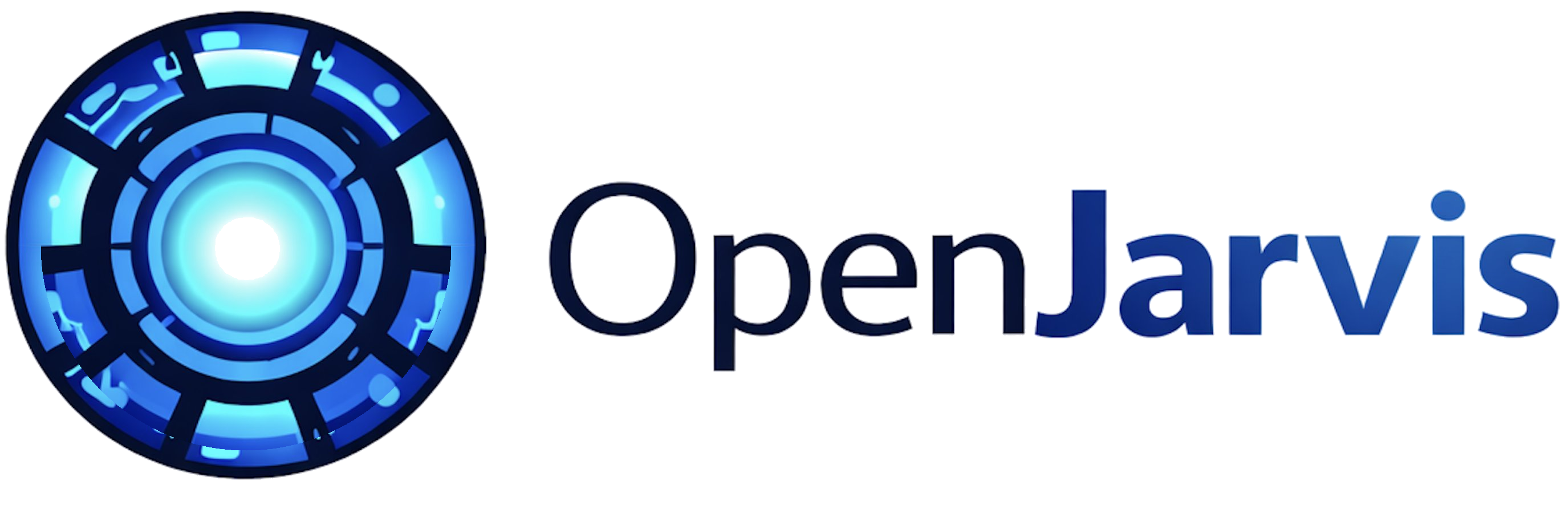}

\author{%
  Jon~Saad-Falcon$^{*1}$ \And
  Avanika~Narayan$^{*1}$ \And
  Robby~Manihani$^{1}$ \\
  Tanvir~Bhathal$^{1}$ \And
  Herumb~Shandilya$^{1}$ \And
  Hakki~Orhun~Akengin$^{1}$ \\
  Gabriel~Bo$^{1}$ \And
  Andrew~Park$^{1}$ \And
  Matthew~Hart$^{1}$ \And
  Caia~Costello$^{2}$ \\
  Chuan~Li$^{2}$ \And
  Christopher~R\'e$^{1}$ \And
  Azalia~Mirhoseini$^{1}$%
}
\affiliations{%
  $^{*}$\,Equal contribution. \quad
  $^{1}$\,Stanford University \quad
  $^{2}$\,Lambda Labs%
}

\date{}
\code{https://github.com/openjarvis/openjarvis}
\website{https://open-jarvis.github.io/OpenJarvis/}

\heroabstract{%
Personal AI stacks, like OpenClaw and Hermes Agent, are becoming central to daily work, yet they route nearly every query (often over sensitive local data) to cloud-hosted frontier models.
Replacing frontier models with local models inside existing stacks does not work: swapping \claudeOpus for Qwen3.5-9B drops accuracy by 25--39~pp across personal AI tasks like PinchBench and GAIA.
Existing stacks bundle agentic prompts, tool descriptions, memory configuration, and runtime settings around a specific cloud model. Only the prompts can be tuned, and state-of-the-art prompt optimizers close just 5~pp of the local--cloud gap on their own.
This motivates a decomposed personal AI stack: one that exposes individual primitives which can be optimized individually or jointly to close the local--cloud gap.
We present \sysname, an architecture that represents a personal AI system as a typed \ojSpec over five primitives: Intelligence, Engine, Agents, Tools \& Memory, and Learning. 
Each primitive is an independently editable field, making the stack end-to-end optimizable and measurable against accuracy, cost, and latency.
Towards closing the local--cloud gap without surrendering local-model properties, \sysname introduces \emph{\methodname}, a local--cloud collaboration in which frontier cloud models propose edits across the \ojSpec at search time, only non-regressing edits are accepted, and the resulting spec runs entirely on-device at inference time.
With \methodname, on-device \ojSpecs match or exceed cloud accuracy on 4 of 8 benchmarks and land within 3.2~pp of the best cloud baseline on average. 
They also reduce marginal API cost by $\sim$800$\times$ and end-to-end latency by 4$\times$.
}

\begin{document}
 
\maketitle

\section{Introduction}
\label{sec:intro}

Personal AI stacks, like OpenClaw~\cite{steinberger2025openclaw} and Hermes Agent~\cite{nousresearch2025hermes}, are now central to daily writing, research, coding, and scheduling. Yet today's personal AI is rarely private or local. Most systems route each query to a cloud-hosted frontier model, often including sensitive personal data. This design demands thousands of dollars per year in recurring API and subscription spend~\cite{zylo2026saasindex, gartner2026inferencecost}. It exposes private data to third-party servers, requires network connectivity to function, and leaves users without ownership of the models they depend on. 
It also consumes orders of magnitude more energy per token than local execution~\cite{saadfalcon2026intelligencewattmeasuringintelligence}. 
Meanwhile, consumer accelerators~\cite{apple2024m4, qualcomm2024hexagon, intel2023meteorlake} now run 1B--128B open-weight models with FP8 quantization~\cite{gerganov2023llamacpp}, and open-weight families such as Qwen3.5~\cite{qwen3.5} and Gemma4~\cite{gemma4_2026} trail frontier cloud models by 6--12 months on personal AI tasks rather than years~\cite{lmsys2024chatbotarena, saadfalcon2026intelligencewattmeasuringintelligence}. 
On-device models nonetheless remain confined to trivial tasks like tone adjustment and text completion~\cite{apple2024foundation}.

In this work we ask: \emph{can the core of a personal AI stack (i.e., model inference, agent execution, memory, and learning) run on-device while remaining competitive with cloud-only stacks?}
We study this question across eight personal AI benchmarks spanning writing, research, coding, and scheduling, including PinchBench~\cite{pinchbench2025} and GAIA~\cite{mialon2024gaia} (Section~\ref{sec:setup}).

\begin{figure*}[t]
    \centering
    \includegraphics[width=\textwidth]{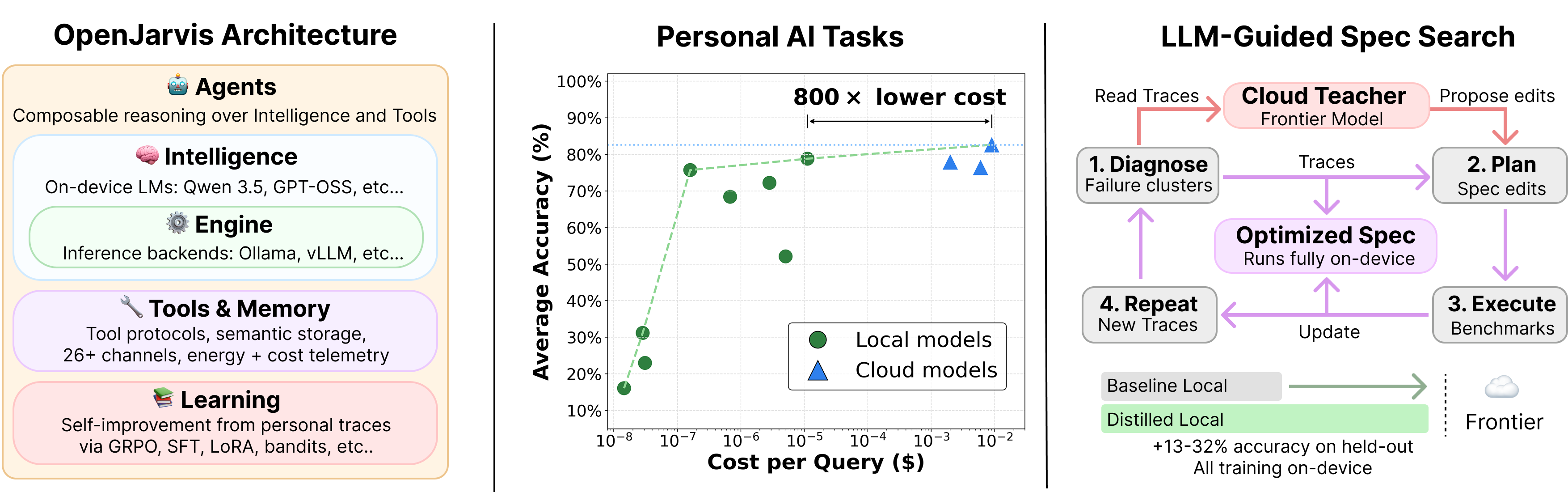}
    \caption{\textbf{Overview of \sysname.}
    \textbf{(Left)}~Five composable primitives (Intelligence, Engine, Agents, Tools \& Memory, Learning) are composed through a declarative \ojSpec that can be shared, evaluated, and optimized end-to-end (Section~\ref{sec:primitives}).
    \textbf{(Middle)}~Joint accuracy-efficiency evaluation across the evaluated local \ojSpecs (green) and 3 cloud baselines (blue) reveals that on-device configurations approach within 3.2~pp of the best cloud model at roughly 800$\times$ lower marginal API cost per query (Section~\ref{sec:efficiency}).
    \textbf{(Right)}~\methodname uses frontier cloud models as reflective proposers: the teacher diagnoses failures, proposes edits across the full \ojSpec, and a held-out gate accepts only non-regressing improvements, closing the remaining local--cloud accuracy gap by 13--32~pp with student training on-device (Section~\ref{sec:distillation_results}).}
    \label{fig:openjarvis_overview}
\end{figure*}

As a first attempt, we ask whether one can simply swap the cloud model for a local one inside an existing stack. Replacing \claudeOpus with \texttt{Qwen3.5-9B}, while keeping the rest of OpenClaw or Hermes Agent fixed, drops accuracy by 25--39 percentage points across PinchBench and GAIA (Section~\ref{sec:portability}, Table~\ref{tab:portability}). We identify two reasons for the local-cloud performance gap:

\begin{itemize}[leftmargin=*]
    \item \emph{Substitution breaks monolithic stacks.} Agent prompts, tool descriptions, memory configuration, and runtime settings are fused into the framework, co-designed for the intended cloud model. Substituting a local model breaks all of these at once (Section~\ref{sec:portability}).
    \item \emph{Single-primitive optimization plateaus.} Of the bundled components, only prompts can be tuned without source-level changes. The rest are baked into the framework. 
    Applying state-of-the-art prompt optimizers (GEPA~\cite{agrawal2026gepareflectivepromptevolution} and DSPy~\cite{khattab2024dspy}) to the swapped-in local model, while holding everything else fixed, closes only 5~pp of the cloud--local gap on their own. Optimizing one primitive at a time cannot deliver the coordinated changes across model, prompts, tools, and runtime that misalignment demands.
\end{itemize}

Motivated by these limitations, we introduce \sysname, an architecture that makes personal AI stacks end-to-end optimizable. \sysname has two core components. 
First, the \emph{spec}: a typed configuration object that decomposes the personal AI system as five editable primitives: 
\emph{Intelligence}, the language model architecture and its weights;
\emph{Engine}, the inference runtime; 
\emph{Agents}, the reasoning loop; 
\emph{Tools \& Memory}, data integration and persistent storage; 
and \emph{Learning}, the optimizer that updates the system from traces (Section~\ref{sec:primitives}). 
The spec exposes every component of the stack as a degree of freedom in one optimizable object. 
Second, \methodname: a local--cloud search algorithm that jointly optimizes all four editable primitives of the spec: Intelligence, Engine, Agents, and Tools \& Memory. 
\methodname runs entirely on-device at inference time but uses frontier cloud models at search time to improve the local spec as much as possible. 
This gives us the best of both worlds: cloud-model capability transferred into a spec that runs locally. 
At each search step, a frontier model reads traces from the current spec and proposes coordinated edits (i.e., rewrite a tool description, adjust a runtime setting) across the four primitives. 
Each edit is accepted only if it improves performance on a held-out evaluation, and accepted edits become the new spec for the next step.

Across eight benchmarks for personal AI, optimized specs match or exceed cloud accuracy on four, land within 3.2~pp of the best cloud baseline on average, and reduce marginal API cost by ${\sim}800\times$ and end-to-end latency by $4\times$ (Section~\ref{sec:distillation_results}). 
\methodname closes 13--32~pp of the cloud--local gap at 7--11$\times$ lower optimization cost than single-primitive baselines. 
We perform a detailed analysis of \methodname's search space. Our analysis highlights three axes that govern the local--cloud gap:

\begin{enumerate}[leftmargin=*]
    \item[(a)] \textbf{Editable surface.} \emph{How does expanding the move space from prompts alone to all four editable primitives affect gap closure?}
    In Figure~\ref{fig:editable_set_ablation} and Appendix~\ref{app:edit_type_ablation}, we show that expanding the editable set adds 5.5--16.5~pp.
    \item[(b)] \textbf{Proposer choice.} \emph{Does diagnosing failures before proposing edits matter, or would evolutionary spec search at the same move space suffice?} In Figure~\ref{fig:editable_set_ablation} and Appendix~\ref{app:proposer_ablation}, we show that the diagnose-and-propose loop in \methodname adds 10.0~pp on average over evolutionary spec search at the same four-primitive move space.
    \item[(c)] \textbf{Search budget.} \emph{How does optimization cost trade off against gap closure?}
    In Figure~\ref{fig:accuracy_vs_optimization_cost}, we show that \methodname matches single-primitive baselines at 7--11$\times$ lower optimization cost.
\end{enumerate}

To summarize, our main contributions are as follows:

\begin{itemize}[leftmargin=*]
    \item Propose \sysname, an architecture that decomposes a personal AI system into five typed primitives composed into an editable spec, making the stack end-to-end optimizable and measurable on accuracy, cost, and latency (Section~\ref{sec:methods}).
    \item Propose \methodname, a local--cloud collaboration that uses frontier models to propose spec edits at search time and runs the resulting spec entirely on-device at inference time, closing 13--32~pp of the cloud--local gap at 7--11$\times$ lower optimization cost than single-primitive baselines (Section~\ref{sec:experiments}).
    \item Conduct an in-depth analysis across eight personal AI benchmarks, showing that on-device specs are Pareto-optimal on marginal API cost and latency with $\sim$800$\times$ lower marginal API cost and 4$\times$ lower end-to-end latency, and isolating the contributions of the proposer and the move space to the overall gain (Section~\ref{sec:efficiency}).
\end{itemize}

\section{Related Work}
\label{sec:related_work}

Unifying abstractions have historically accelerated ML progress. 
Imperative deep-learning frameworks~\cite{paszke2019pytorch} and type-based declarative ML toolboxes~\cite{molino2019ludwig} unified model construction, while data systems for monitoring deployed ML products~\cite{re2019overton} and LM pipeline compilers~\cite{khattab2024dspy} unified iterative improvement.
The personal AI ecosystem is undergoing a similar fragmentation-to-unification transition.
Existing tools each formalize a subset of the five primitives (Intelligence, Engine, Agents, Tools \& Memory, and Learning) while hardcoding or omitting the rest.

\paragraph{Agents and Tools: personal AI stacks.}
Existing personal AI stacks formalize Agents and Tools as configurable layers but tie Intelligence, Engine, and Learning to specific cloud models, so substituting a different Intelligence degrades the full pipeline (Section~\ref{sec:portability}).
This pattern holds across open frameworks such as OpenClaw~\cite{steinberger2025openclaw}, Hermes Agent~\cite{nousresearch2025hermes}, LangChain~\cite{chase2022langchain}, CrewAI~\cite{crewai2024}, Google ADK~\cite{google2025adk}, OpenAI Symphony~\cite{openai2025symphony}, Qwen-Agent~\cite{qwenagent2025}, and related systems~\cite{picoclaw2025, edgeclaw2025, nanobot2025, ironclaw2025, tinyclaw2025, zeroclaw2025, mimiclaw2025}, whose Agent prompts, Tool descriptions, Memory configuration, and Engine/runtime settings are tied to specific model and engine choices.
Closed systems (Apple Intelligence~\cite{apple2024foundation}, Gemini Nano~\cite{gemini_nano_android}) additionally hardcode the Engine to proprietary runtimes and expose no Learning interface.

\paragraph{Engine and Intelligence: local inference tools.}
Local inference tools formalize Engine and Intelligence as configurable layers but provide no Agents, Tools, or Learning, leaving the upper half of the stack unaddressed.
This includes serving runtimes (Ollama~\cite{ollama2023}, llama.cpp~\cite{gerganov2023llamacpp}, LM Studio~\cite{lmstudio2024}, LocalAI~\cite{localai2024}, gemma.cpp~\cite{gemmacpp2024}), Engine-side performance optimizations such as quantization (GPTQ~\cite{frantar2023gptq}, AWQ~\cite{lin2024awq}), speculative decoding~\cite{leviathan2023speculative, chen2023accelerating}, and hardware-aware serving (MLC-LLM~\cite{mlcllm2023}, ExecuTorch~\cite{executorch2024}, vLLM~\cite{kwon2023vllm}, SGLang~\cite{zheng2024sglang}), and Intelligence-side on-device architectures (MobileLLM~\cite{liu2024mobilellm}, Gemma~3n~\cite{gemma3n2025}, Liquid Nanos~\cite{liquidnanos2025}, Nemotron-Flash~\cite{nemotronflash2025}) co-designed for efficient execution.
\sysname's Engine abstracts over these backends with built-in energy and cost telemetry, extending learned query routing~\cite{llmrouter2025} to local-vs-cloud decisions.

\paragraph{Learning: optimization methods.}
Optimization methods formalize Learning but address one primitive at a time, falling into two families.
Weight optimizers update Intelligence: classical knowledge distillation~\cite{hinton2015distilling} (see~\cite{xu2024kd_survey} for a survey) trains students to match teacher soft targets, MiniLLM~\cite{gu2024minillm} and DeepSeek-R1~\cite{deepseek2025r1} extend this to LLMs, and PEFT methods (LoRA~\cite{hu2022lora}, QLoRA~\cite{dettmers2023qlora}, GRPO~\cite{shao2024deepseekmath}) make on-device weight updates feasible even on microcontrollers~\cite{lin2022ondevice256kb}, with cross-platform mobile LoRA now practical~\cite{tetherdata2025qvac, mobilefinetuner2025}.
Prompt and agent optimizers update Agents (and, in some variants, Tools): DSPy~\cite{khattab2024dspy} and ACE~\cite{zhang2026agenticcontextengineeringevolving} compile or evolve agent prompts, and GEPA~\cite{agrawal2026gepareflectivepromptevolution} uses LLM-based reflection over trajectories with Pareto-efficient evolutionary machinery to preserve and merge complementary candidates.
Inference-time collaboration (Minions~\cite{narayan2025minions}, Advisor Models~\cite{asawa2025advisor}) decomposes tasks across local and cloud Intelligence but does not persistently update any primitive, and OpenClaw-RL~\cite{wang2026openclawrl} is the closest prior system to continuous on-device learning, training an RL policy from live user interactions but updating only Intelligence weights in a cloud-hosted loop.
Prior work has also observed that combining weight and prompt optimization can outperform either alone~\cite{soylu2024bettertogether}.
\methodname (Section~\ref{sec:distillation}) extends this direction to four editable primitives (Intelligence, Engine, Agents, and Tools \& Memory) with Learning as the optimizer slot, making it complementary to single-primitive methods rather than a replacement for them.

\paragraph{Joint evaluation.}
On the evaluation side, tools exist for individual efficiency dimensions (Zeus~\cite{zeus2023}, AI Energy Score~\cite{aienergyscore2024}, MLCommons~\cite{mlcommons2024}) but measure a single axis of a single primitive.
Standard agent benchmarks (GAIA~\cite{mialon2024gaia}, SWE-bench~\cite{jimenez2024swebench}, Terminal-Bench~\cite{merrill2026terminalbench}) report accuracy alone and assume unlimited cloud compute.
Because the spec provides a common representation for complete configurations, we are not aware of a prior framework that jointly evaluates accuracy, energy, latency, power, and cost across the full five-primitive composition (Section~\ref{sec:experiments}).

\paragraph{Summary.}
Each prior effort formalizes a subset of the structure: personal AI frameworks address Agents and Tools, local inference tools address Engine and Intelligence, optimization methods address Learning one primitive at a time, and evaluation tools measure one axis of one primitive.
No existing framework formalizes the full five-primitive composition or exposes it as a single optimizable object.
This gap motivates the spec abstraction and \methodname we introduce in Section~\ref{sec:methods}.

\section{Methods}
\label{sec:methods}

We present \sysname in three parts: the spec abstraction (Section~\ref{sec:primitives}), joint accuracy-efficiency evaluation (Section~\ref{sec:evaluation}), and \methodname (Section~\ref{sec:distillation}).
The core idea is that every local model needs the surrounding agent system reconfigured around it; the spec makes that reconfiguration explicit and optimizable.
Section~\ref{sec:related_work} situates these design choices in prior work, and Appendix~\ref{app:methods} provides expanded method details.

\subsection{Primitives and the Spec Abstraction}
\label{sec:primitives}

\begin{figure}[t]
    \centering
    \includegraphics[width=\linewidth]{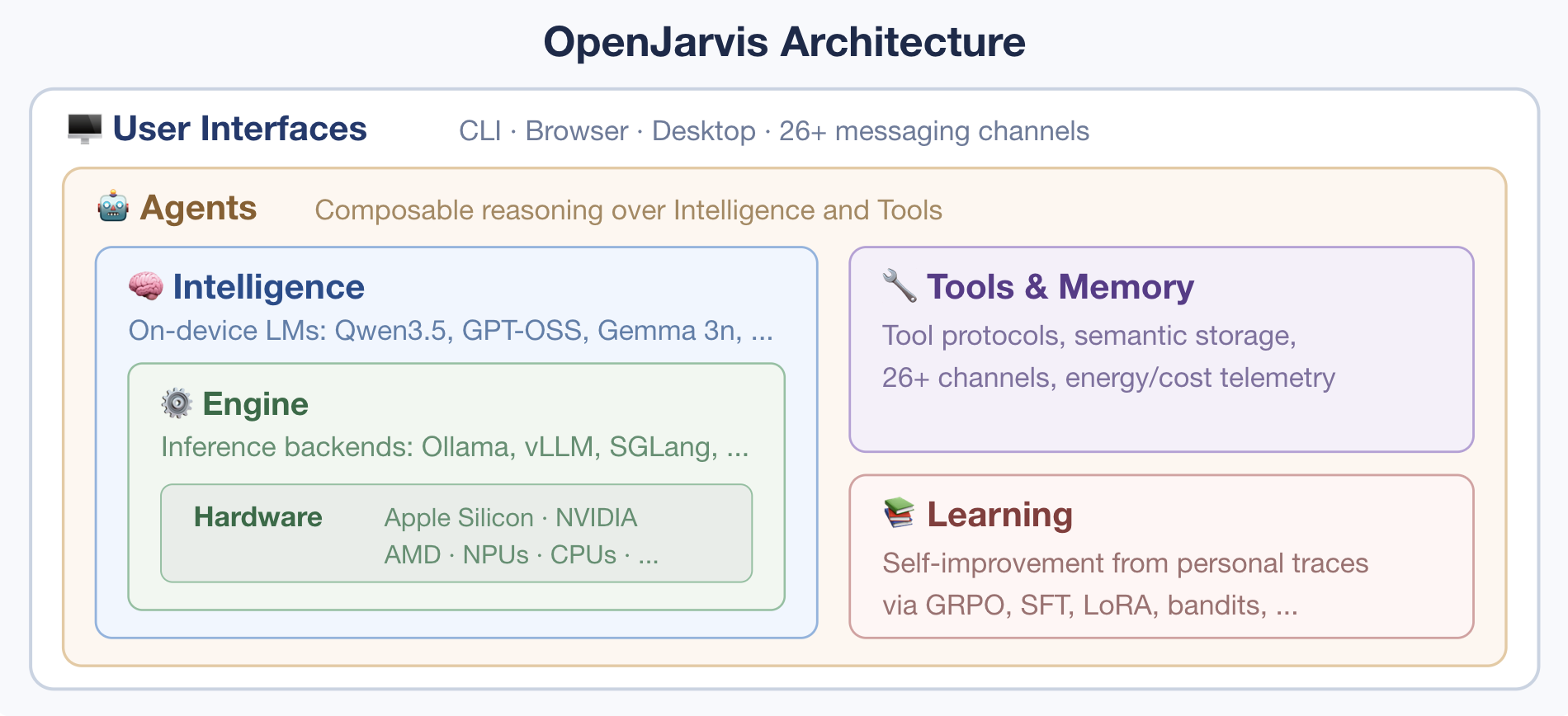}
    \caption{
        \textbf{OpenJarvis architecture.}
        Five composable primitives decouple model selection (\emph{Intelligence}), inference runtime (\emph{Engine}), agent logic (\emph{Agents}), data integration (\emph{Tools \& Memory}), and on-device learning (\emph{Learning}) into independently swappable layers.
        A \emph{spec} (Section~\ref{sec:primitives}) composes all five into a declarative configuration that can be shared, evaluated, and optimized end-to-end.
    }
\end{figure}

\sysname decomposes a personal AI stack into five primitives, each with a typed interface and registry (Figure~\ref{fig:openjarvis_overview}, left).
\emph{Intelligence} specifies the language model and generation parameters.
\emph{Engine} specifies the runtime, hardware execution path, batching, quantization, and cache settings.
\emph{Agents} specify the reasoning loop, prompts, and tool-use policy.
\emph{Tools \& Memory} specify external interfaces, retrieval, and persistent user state.
\emph{Learning} specifies the optimizer that updates the spec from traces, including weight-only optimizers such as LoRA~\cite{hu2022lora}, prompt optimizers such as DSPy~\cite{khattab2024dspy}, and \methodname (Section \ref{sec:distillation}).
Implementation details, supported runtimes, connectors, memory systems, and optimizer variants are in Appendix~\ref{app:primitives}.

\paragraph{Five-primitive architecture.}
Each primitive is an independent degree of freedom that prior systems optimize in isolation.
The decomposition is designed to separate choices that are independently configurable in deployed systems.
For example, Intelligence and Engine separate model selection (\texttt{Qwen3.5-9B} vs.\ \texttt{Qwen3.5-122B}) from runtime selection (\texttt{vLLM} vs.\ \texttt{Ollama}).
Agents and Tools separate the reasoning loop (ReAct vs.\ CodeAct) from the interfaces it invokes (filesystem, web search, MCP servers).
Learning is structurally different from the other four primitives: it is the primitive we use to optimize all the others, making edits to Intelligence weights, Agent prompts/logic, Tools \& Memory selection and descriptions, and Engine runtime specifications.
This primitive lets the same framework describe systems that share Intelligence and Agent configurations but differ only in Learning policy, such as OpenClaw-RL~\cite{wang2026openclawrl}.
Section~\ref{sec:distillation_results} tests whether these distinctions between primitives matter for constructing and optimizing local AI.

\begin{figure}[t]
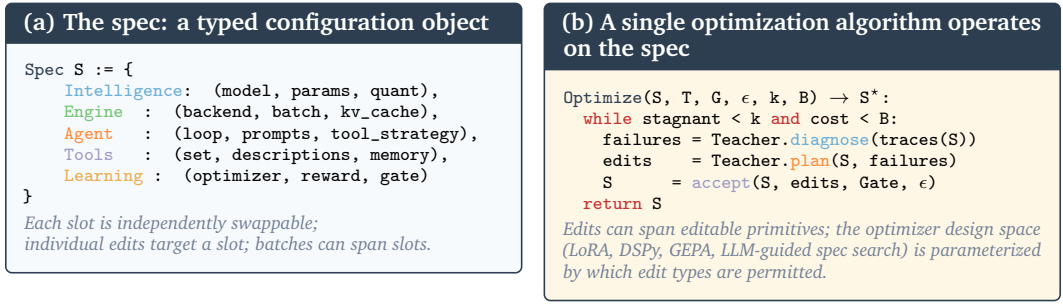

\centering

\definecolor{intelligenceClr}{HTML}{6BAED6}   %
\definecolor{engineClr}{HTML}{74C476}         %
\definecolor{agentClr}{HTML}{FD8D3C}          %
\definecolor{toolsClr}{HTML}{9E9AC8}          %
\definecolor{learningClr}{HTML}{E9AC4D}       %
\definecolor{specBg}{HTML}{F7FAFC}            %
\definecolor{optBg}{HTML}{FFF7E6}             %
\definecolor{frameClr}{HTML}{2C3E50}          %
\definecolor{commentClr}{HTML}{718096}        %
\definecolor{keywordClr}{HTML}{C53030}        %

\begin{minipage}[t]{0.49\linewidth}
\vspace{0pt}
\begin{tcolorbox}[
  colback=specBg, colframe=frameClr, boxrule=0.6pt,
  arc=2pt, left=4pt, right=4pt, top=4pt, bottom=4pt,
  title={\textbf{(a)~The spec: a typed configuration object}},
  coltitle=white, fonttitle=\small\bfseries
]
\scriptsize\ttfamily
\textcolor{frameClr}{\textbf{Spec}}~\textbf{S}~:=~\{\\
\hspace*{2em}\textcolor{intelligenceClr}{\textbf{Intelligence}}: (model, params, quant),\\
\hspace*{2em}\textcolor{engineClr}{\textbf{Engine}}\hspace*{1em}: (backend, batch, kv\_cache),\\
\hspace*{2em}\textcolor{agentClr}{\textbf{Agent}}\hspace*{1.5em}: (loop, prompts, tool\_strategy),\\
\hspace*{2em}\textcolor{toolsClr}{\textbf{Tools}}\hspace*{1.5em}: (set, descriptions, memory),\\
\hspace*{2em}\textcolor{learningClr}{\textbf{Learning}}\hspace*{0.5em}: (optimizer, reward, gate)\\
\}\\[2pt]
\rmfamily\textcolor{commentClr}{\itshape Each slot is independently swappable;\\
individual edits target a slot; batches can span slots.}
\end{tcolorbox}
\end{minipage}\hfill
\begin{minipage}[t]{0.49\linewidth}
\vspace{0pt}
\begin{tcolorbox}[
  colback=optBg, colframe=frameClr, boxrule=0.6pt,
  arc=2pt, left=4pt, right=4pt, top=4pt, bottom=4pt,
  title={\textbf{(b)~A single optimization algorithm operates on the spec}},
  coltitle=white, fonttitle=\small\bfseries
]
\scriptsize\ttfamily
\textcolor{frameClr}{\textbf{Optimize}}(S, T, G, $\epsilon$, k, B) $\to$ S$^\star$:\\
\hspace*{1em}\textcolor{keywordClr}{\textbf{while}} stagnant < k \textcolor{keywordClr}{\textbf{and}} cost < B:\\
\hspace*{2em}failures = Teacher.\textcolor{intelligenceClr}{\textbf{diagnose}}(traces(S))\\
\hspace*{2em}edits\hspace*{1em}\hspace*{0.5em} = Teacher.\textcolor{agentClr}{\textbf{plan}}(S, failures)\\
\hspace*{2em}S\hspace*{2em}\hspace*{0.5em} = \textcolor{toolsClr}{\textbf{accept}}(S, edits, Gate, $\epsilon$)\\
\hspace*{1em}\textcolor{keywordClr}{\textbf{return}} S\\[2pt]
\rmfamily\textcolor{commentClr}{\itshape Edits can span editable primitives; the optimizer design space\\
(LoRA, DSPy, GEPA, \methodname) is parameterized by which edit types are permitted.}
\end{tcolorbox}
\end{minipage}

\caption{\textbf{The spec abstraction.}
\textbf{(a)}~A spec is a typed configuration object with five primitives: Intelligence, Engine, Agents, Tools \& Memory, and Learning.
\textbf{(b)}~Optimizers instantiate the same signature by restricting which fields they edit.
LoRA edits Intelligence weights; DSPy and GEPA edit Agent prompts; \methodname edits Intelligence, Engine, Agents, and Tools \& Memory jointly.}
\label{fig:spec_pseudocode}
\end{figure}

\paragraph{The spec abstraction.}
The five primitives are composed through the \emph{spec}: a typed configuration object on which a single optimization algorithm operates.
Figure~\ref{fig:spec_pseudocode} shows the spec as a structured declaration alongside the optimization signature; Figure~\ref{fig:spec_example} (Appendix~\ref{app:primitives}) shows the corresponding TOML serialization.
A spec can be versioned, shared, evaluated on standardized benchmarks, and optimized end-to-end (Section~\ref{sec:distillation}).

\subsection{Evaluation Metrics}
\label{sec:evaluation}

We evaluate complete \ojSpecs rather than isolated models.
For each spec, an instrumented wrapper records five quantities per query: accuracy, energy, latency, power, and dollar cost.
Accuracy is exact match or LLM-judge win rate, depending on the benchmark.
Energy is measured through vendor APIs for local hardware and estimated for cloud baselines following prior work~\cite{saadfalcon2026intelligencewattmeasuringintelligence}.
Latency is end-to-end wall-clock time, and power is energy divided by latency.
Dollar cost is per-token API pricing and tool fees; local model inference is reported as \$0 marginal API cost, with hardware and electricity reported separately.
Together, the spec and wrapper expose Pareto structure that accuracy-only benchmarks and single-axis efficiency tools miss~\cite{mialon2024gaia, jimenez2024swebench, merrill2026terminalbench, zeus2023, aienergyscore2024, mlcommons2024}.

\subsection{\methodname}
\label{sec:distillation}

\methodname optimizes a spec by searching over its primitive fields, where each \ojSpec is one complete local AI configuration and changing its fields defines the search space.
The Learning primitive specifies how a spec is updated from traces, namely which edits are considered, how edited specs are evaluated, and when optimization stops.
In \methodname, this loop is a local--cloud collaboration: frontier cloud models, which excel at reading traces and reasoning about coordinated edits across many primitives, propose changes to the spec at search time, while local hardware, which excels at running the resulting configuration with low latency and zero marginal API cost, executes the optimized spec at inference time.
A frontier model reads eligible traces, identifies failure patterns, and proposes candidate edits across the editable primitives.

\paragraph{Failure clusters and the gate.}
The frontier model groups traces sharing a common failure mode into \emph{failure clusters}, each annotated with student vs.\ teacher success rates and a natural-language characterization of the skill gap (e.g., ``student fails on multi-hop questions requiring calendar lookups because it does not invoke the calendar tool''); see Appendix~\ref{app:diagnose} for the full clustering protocol.
Each candidate edit is evaluated on held-out examples for the task, and accepted only if it improves the targeted failure cluster without causing unacceptable regressions elsewhere.
We refer to this held-out validation check as the \emph{gate}.
Intelligence edits change model parameters or training triggers; Engine edits change runtime and serving choices; Agent edits change prompts, reasoning loops, and verification; Tools \& Memory edits change Tool availability, Tool descriptions, and Memory configuration.

\paragraph{What runs where.}
At inference time, the resulting spec runs on-device for model inference and agent execution, and when an Intelligence edit triggers student training, that training also runs locally.
Teacher calls provide diagnoses, edit proposals, and labels, not inference-time model calls.
Cloud-as-tool use is disabled in the headline local configurations.

\paragraph{Relation to single-primitive optimizers.}
Most existing optimizers update one primitive at a time, falling into two families.
Prompt and agent optimizers such as DSPy~\cite{khattab2024dspy}, GEPA~\cite{agrawal2026gepareflectivepromptevolution}, and ACE~\cite{zhang2026agenticcontextengineeringevolving} edit Agents and, in some cases, Tools, while weight optimizers such as SFT, LoRA~\cite{hu2022lora}, and GRPO~\cite{shao2024deepseekmath} edit Intelligence.
\methodname is complementary rather than competing: it operates on the same spec these methods would edit, and a single proposal can change any combination of the four editable primitives at once.
The defining property of the spec is that Intelligence, Engine, Agents, and Tools \& Memory fields are degrees of freedom in one optimizable object, so a single spec edit can simultaneously rewrite a Tool description and update model weights against the resulting prompt, something neither family of single-primitive optimizers expresses natively.
Figure~\ref{fig:search_regimes} summarizes the three regimes compared in Section~\ref{sec:distillation_results}: \methodname (Algorithm~\ref{alg:spec_distillation}), an evolutionary spec-search baseline (Algorithm~\ref{alg:gepa_genetic}), and single-component optimization (Algorithm~\ref{alg:single_component}); Figure~\ref{fig:editable_set_ablation} separates the proposer axis from the move-space axis.

\begin{figure*}[t]
\centering
\definecolor{algBgLeft}{HTML}{F0F7FB}
\definecolor{algBgMid}{HTML}{F7F5FF}
\definecolor{algBgRight}{HTML}{FAF0F0}
\definecolor{frameClr}{HTML}{2C3E50}
\definecolor{commentClr}{HTML}{C53030}
\definecolor{primitiveClr}{HTML}{6B46C1}

\refstepcounter{algorithm}\label{alg:spec_distillation}%
\begin{tcolorbox}[
  colback=algBgLeft, colframe=frameClr, boxrule=0.6pt,
  arc=2pt, left=4pt, right=4pt, top=2pt, bottom=2pt,
  title={\textbf{Algorithm~\thealgorithm.~\methodname: greedy gated edits across primitives}},
  coltitle=white, fonttitle=\footnotesize\bfseries
]
\scriptsize
\renewcommand{\algorithmicindent}{0.8em}
\begin{algorithmic}[1]
\Require Spec $S_0$, teacher $T$, gate $G$, regression tolerance $\epsilon$, budget $B$
\State $S \gets S_0$
\While{not converged \textbf{and} cost $< B$}
  \State $C \gets T.\mathrm{diagnose}(\mathrm{traces}(S))$
  \State $e \gets T.\mathrm{propose}(S, C)$ \Comment{may edit Intelligence, Engine, Agents, and Tools \& Memory jointly}
  \State $S' \gets \mathrm{apply}(S,e)$
  \If{$\mathrm{GateOK}(S',S,C,\epsilon)$}
    \State $S \gets S'$ \Comment{greedy accept}
  \EndIf
\EndWhile
\State \Return $S$
\end{algorithmic}
\vspace{1pt}
\footnotesize\textcolor{commentClr}{\textit{$\mathrm{GateOK}$ means the target cluster improves and every non-target cluster regresses by at most $\epsilon$.}}
\end{tcolorbox}

\vspace{4pt}
\begin{minipage}[t]{0.49\textwidth}
\vspace{0pt}%
\refstepcounter{algorithm}\label{alg:gepa_genetic}%
\begin{tcolorbox}[
  colback=algBgMid, colframe=frameClr, boxrule=0.6pt,
  arc=2pt, left=4pt, right=4pt, top=2pt, bottom=2pt,
  title={\textbf{Algorithm~\thealgorithm.~Evolutionary spec search}},
  coltitle=white, fonttitle=\footnotesize\bfseries
]
\scriptsize
\renewcommand{\algorithmicindent}{0.8em}
\begin{algorithmic}[1]
\Require Initial candidate $S_0$, reflection LM $T$, evaluator $G$, budget $B$
\State Initialize population $\mathcal{P} \gets \{S_0\}$
\While{cost $< B$}
  \State $S \gets$ sample candidate from Pareto frontier of $\mathcal{P}$
  \State $e \gets T.\mathrm{reflect}(S,\mathrm{traces}(S))$
  \State $S' \gets \mathrm{apply}(S,e)$ \Comment{reflective mutation}
  \State Optionally merge $S'$ with frontier candidate $S_j$
  \State Evaluate candidates and update Pareto frontier $\mathcal{P}$
\EndWhile
\State \Return best candidate in $\mathcal{P}$
\end{algorithmic}
\end{tcolorbox}
\end{minipage}\hfill
\begin{minipage}[t]{0.49\textwidth}
\vspace{0pt}%
\refstepcounter{algorithm}\label{alg:single_component}%
\begin{tcolorbox}[
  colback=algBgRight, colframe=frameClr, boxrule=0.6pt,
  arc=2pt, left=4pt, right=4pt, top=2pt, bottom=2pt,
  title={\textbf{Algorithm~\thealgorithm.~Single-component optimizer}},
  coltitle=white, fonttitle=\footnotesize\bfseries
]
\scriptsize
\renewcommand{\algorithmicindent}{0.8em}
\begin{algorithmic}[1]
\Require Spec $S_0$, dataset $\mathcal{D}$, edit type $\tau \in \{\mathrm{Intelligence},\mathrm{Agent}\}$, budget $B$
\State $S \gets S_0$
\While{cost $< B$}
  \State $e \gets$ propose edit of type $\tau$ \Comment{restricted to one primitive}
  \State $S' \gets \mathrm{apply}(S,e)$
  \If{loss or validation score improves on $\mathcal{D}$}
    \State $S \gets S'$
  \EndIf
\EndWhile
\State \Return $S$
\end{algorithmic}
\vspace{2pt}
\footnotesize\textcolor{commentClr}{\textit{Cannot exploit coupling across primitives.}}
\end{tcolorbox}
\end{minipage}

\caption{\textbf{Three ways to optimize a spec.}
\sysname proposes edits to the four editable primitives and keeps an edit only if held-out performance does not regress.
Evolutionary spec search maintains and merges a population of candidate \ojSpecs \cite{agrawal2026gepareflectivepromptevolution}.
Single-component baselines edit one primitive at a time.}
\label{fig:search_regimes}
\end{figure*}

\paragraph{Search loop.}
Each session repeats four steps: diagnose failures, propose edits, execute candidates, and commit only gated improvements.
Diagnose clusters traces where the student underperforms.
Plan proposes edits over Intelligence, Engine, Agents, and Tools \& Memory fields.
Execute evaluates each edited spec on a held-out gate built from eligible traces, agentic datasets with known answers~\cite{turner2025generalthought, su2025toolorchestra}, and standard benchmark splits~\cite{wang2024mmlupro, mialon2024gaia, yao2024taubench}.
Repeat commits accepted edits and rolls back rejected ones.
The loop stops when gate scores stagnate for $k$ sessions (default $k{=}5$) or the budget is exhausted.
Full per-phase examples are in Appendix~\ref{app:distillation}.

\paragraph{Gate score.}
Let $G(S)$ be the held-out gate score of spec $S$, and let $G_c(S)$ be the score restricted to failure cluster $c$.
For an edit $e$ targeting cluster $c$, with $S'=\mathrm{apply}(S,e)$, we accept if and only if:
\[
G_c(S') > G_c(S)
\quad\text{and}\quad
G_{c'}(S') \geq G_{c'}(S)-\epsilon
\quad \forall c'\neq c .
\]
The default tolerance is $\epsilon=1\%$.
This single rule applies across Intelligence weights, Agent prompts, Tool descriptions, Memory configuration, and Engine/runtime settings.

\paragraph{(Optional) Composite reward for Intelligence edits.}
If Intelligence edits are requested involving model training, each candidate response $y$ to a query $q$ is scored by a composite reward over accuracy, energy, latency, and cost:
\begin{equation}
    R(q, y) = \alpha R_\text{acc}(q, y) - \beta \hat{E}(q, y) - \gamma \hat{L}(q, y) - \delta \hat{C}(q, y).
    \label{eq:reward}
\end{equation}
The default weights are $(\alpha,\beta,\gamma,\delta)=(0.5,0.1,0.1,0.3)$.
The reward scores responses during GRPO training within an Intelligence edit.
The gate evaluates the resulting spec as a whole.
Normalization details and robustness checks are in Appendix~\ref{app:distillation_robustness}.

\section{Experiments}
\label{sec:experiments}

\subsection{Models, Benchmarks, and Hardware}
\label{sec:setup}

We evaluate \sysname across 8 benchmarks, 11 local models from 4 families, 3 cloud baselines, and 7 hardware platforms.
The suite contains 508 tasks, and agentic benchmarks enforce a 2-hour timeout per task.
All results report the mean over 5 independent runs.
Scoring uses \texttt{GPT-5-mini} as judge except where benchmarks provide deterministic grading.
Appendix~\ref{app:experiments} details benchmarks (Appendix~\ref{app:benchmarks}), model choices and full accuracy tables (Appendices~\ref{app:models} and~\ref{app:full_accuracy_table}), hardware (Appendix~\ref{app:hardware}), and protocol (Appendix~\ref{app:protocol}).

\subsection{The Spec Recovers 56--77\% of the Drop and Lands Within 3.2 pp of Cloud on Average}
\label{sec:portability}

The spec abstraction (Section~\ref{sec:primitives}) restores most of the portability that is lost when local models are dropped into cloud-designed frameworks, and the resulting on-device specs land within 3.2~pp of the best cloud model on average.
We establish this with a controlled portability experiment on two production frameworks (Table~\ref{tab:portability}) and a local-vs-cloud accuracy frontier drawn from the full accuracy table in Appendix~\ref{app:full_accuracy_table}.

\begin{table}[ht]
\centering
\small
\setlength{\tabcolsep}{2.5pt}
\renewcommand{\arraystretch}{0.95}
\begin{tabular}{@{}l l cc cc@{}}
\toprule
& & \multicolumn{2}{c}{\texttt{PinchBench}} & \multicolumn{2}{c}{\texttt{GAIA}} \\
\cmidrule(lr){3-4} \cmidrule(lr){5-6}
\textbf{Framework} & \textbf{Condition} & Acc.\ (\%) & $\Delta$ vs.\ (a) & Acc.\ (\%) & $\Delta$ vs.\ (a) \\
\midrule
\multirow{3}{*}{OpenClaw~\cite{steinberger2025openclaw}}
& (a) Default + cloud model     & 96.0 & Ref.      & 58.0 & Ref.      \\
& (b) Default + Qwen3.5-9B      & 62.3 & $-$33.7   & 19.2 & $-$38.8   \\
& (c) \sysname(with Qwen3.5-9B) & 88.4 & $-$7.6 & 41.5 & $-$16.5 \\
\midrule
\multirow{3}{*}{Hermes Agent~\cite{nousresearch2025hermes}}
& (a) Default + cloud model     & 93.5 & Ref.      & 55.1 & Ref.      \\
& (b) Default + Qwen3.5-9B      & 68.7 & $-$24.8   & 22.4 & $-$32.7   \\
& (c) \sysname(with Qwen3.5-9B) & 87.9 & $-$5.6 & 40.8 & $-$14.3 \\
\midrule
\multicolumn{2}{l}{\emph{Recovery} (a)$\to$(b) drop closed by (c)} & \multicolumn{2}{c}{77\% of PB drop} & \multicolumn{2}{c}{57\% of GAIA drop} \\
\bottomrule
\end{tabular}
\caption{\textbf{Portability triangulation: cloud-designed frameworks collapse when local models are substituted; \sysname recovers most of the drop.}
Three conditions per framework: (a)~default + intended cloud model, (b)~default + \texttt{Qwen3.5-9B}, (c)~\sysname spec with Intelligence fixed at \texttt{Qwen3.5-9B} (only Engine, Agent, and Tool \& Memory fields optimized). All $\Delta$ values are reported relative to the default cloud configuration (a), so (b) shows the full local-substitution drop and (c) shows the residual gap to cloud after the spec is retargeted at constant Intelligence. \emph{Recovery} is the fraction of the (a)$\to$(b) drop closed by (c), isolating the contribution of the spec at constant Intelligence. The cloud model is \claudeOpus for both frameworks. Results are averaged over 5 independent runs.}
\label{tab:portability}
\begin{minipage}{\linewidth}
\end{minipage}
\end{table}

\paragraph{\sysname reduces the local-substitution drop from 25--39 pp to 5.6--16.5 pp.}
Table~\ref{tab:portability} isolates portability at fixed Intelligence.
Replacing each framework's intended cloud model with \texttt{Qwen3.5-9B} drops accuracy by 24.8--38.8 pp.
With the same local model under an \sysname spec, the remaining drop relative to the cloud framework is only 5.6--16.5 pp.
Equivalently, the spec closes 77\% of the \texttt{PinchBench} drop and 56--57\% of the \texttt{GAIA} drop.
The drop is architectural rather than capability-bound: OpenClaw and Hermes Agent package Agent prompts, Tool descriptions, Memory configuration, Engine/runtime settings, and model-specific output expectations into personal AI stacks tuned to their intended cloud models.
The \sysname \ojSpec exposes these assumptions as typed slots.
Engine, Agent, and Tool fields can then be retargeted while holding Intelligence fixed.
\texttt{GAIA} recovers less because its deep-reasoning demands exceed what a 9B model can deliver, regardless of configuration.

\paragraph{The best local model lands within 3.2 pp of the best cloud model on average and matches or exceeds cloud on 4 of 8 benchmarks.}
Across the evaluated local \ojSpecs, the best single on-device spec, \texttt{Qwen3.5-122B}, reaches 80.3\% average accuracy, within 3.2 pp of \claudeOpus at 83.5\% (Appendix~\ref{app:full_accuracy_table}, Table~\ref{tab:main_table_results}).
At the per-benchmark level, the best local model on each benchmark matches or exceeds the best cloud model on 4 of 8 benchmarks: \texttt{ToolCall-15}, \texttt{PinchBench}, \texttt{LiveCodeBench}, and $\tau$-Bench~V2.
The remaining gaps concentrate on \texttt{GAIA}, $\tau^2$-Bench~Telecom, and \texttt{DeepResearchBench}, motivating the search experiments in Section~\ref{sec:distillation_results}.

\subsection{Local Models Trade 3.2 pp Accuracy for 800\texorpdfstring{$\times$}{x} Lower Cost and 4\texorpdfstring{$\times$}{x} Lower Latency}
\label{sec:efficiency}

Local configurations form the accuracy--cost Pareto frontier: they match cloud accuracy on four of eight benchmarks while reducing marginal API cost by roughly 800$\times$ and end-to-end latency by roughly 4$\times$.
This matters because heavy users of cloud personal AI spend thousands per year on one assistant, and none of that spend buys ownership, offline access, or price stability.
We establish the comparison by treating the spec as the unit of analysis and instrumenting every query (Section~\ref{sec:evaluation}), so local and cloud configurations can be compared on the same axes.

\begin{figure}[t]
    \centering
    \includegraphics[width=\columnwidth]{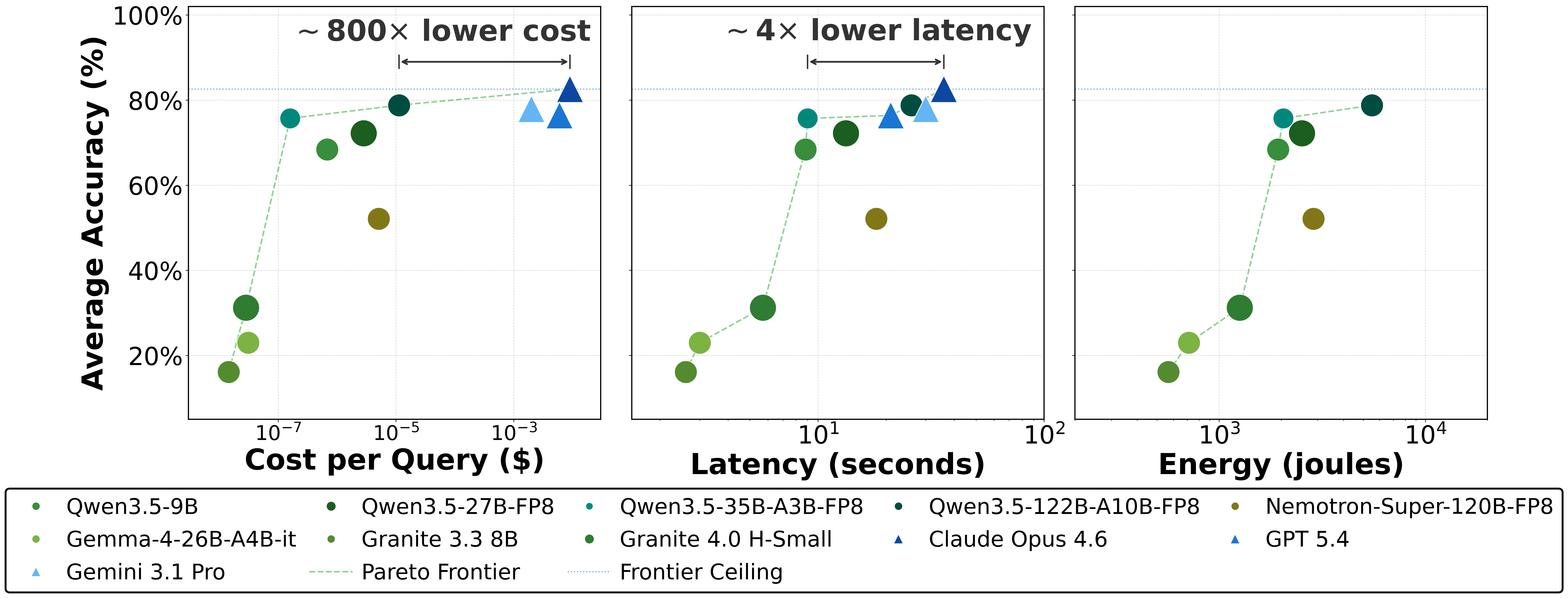}
    \caption{\textbf{Accuracy-efficiency frontier.}
Local configurations approach the best cloud accuracy within 3.2 pp while reducing marginal API cost by roughly 800$\times$ and end-to-end latency by roughly 4$\times$ under our benchmark protocol.
Energy and hardware-specific breakdowns are in Appendix~\ref{app:hardware_profiling}.}
    \label{fig:accuracy_vs_efficiency_graph}
\end{figure}

\paragraph{Local configurations define the accuracy and cost frontier.}
Figure~\ref{fig:accuracy_vs_efficiency_graph} plots average accuracy against dollar cost, latency, and energy per query for the evaluated local \ojSpecs and 3 cloud baselines.
Across cost and latency, the non-dominated configurations are local.
\texttt{Qwen3.5-122B} reaches 80.3\% average accuracy at roughly a thousandth of a cent per query, versus \$0.009 for \claudeOpus at 83.5\%, giving an approximately 800$\times$ marginal API-cost advantage for a 3.2 pp accuracy deficit.
The cost axis reports API fees only; hardware and electricity are accounted for separately via measured energy and amortization (Appendix~\ref{app:hardware_profiling}).
Local configurations also complete the full agentic workloads roughly 4$\times$ faster in our protocol, though single-shot prompts can favor cloud serving due to time-to-first-token optimizations. 
The resulting Pareto structure motivates spec-level model and Engine selection: for a given accuracy target, a \ojSpec can select the local configuration that minimizes cost, latency, or energy.

\subsection{\methodname Shrinks the Cloud--Local Gap by 13--32 pp at 7--11\texorpdfstring{$\times$}{x} Lower Cost}
\label{sec:distillation_results}

Sections~\ref{sec:portability} and~\ref{sec:efficiency} show that on-device \ojSpecs can match cloud on 4 of 8 benchmarks, but gaps remain on reasoning- and research-heavy tasks.
We evaluate whether \methodname closes this gap on \texttt{PinchBench}, \texttt{LiveCodeBench}, and \texttt{LiveResearchBench}.
For each benchmark, we run search with four local target models and three frontier proposer models; Appendix~\ref{app:full_distillation} extends the evaluation to all 8 benchmarks.

\begin{figure}[t]
    \centering
    \includegraphics[width=\columnwidth]{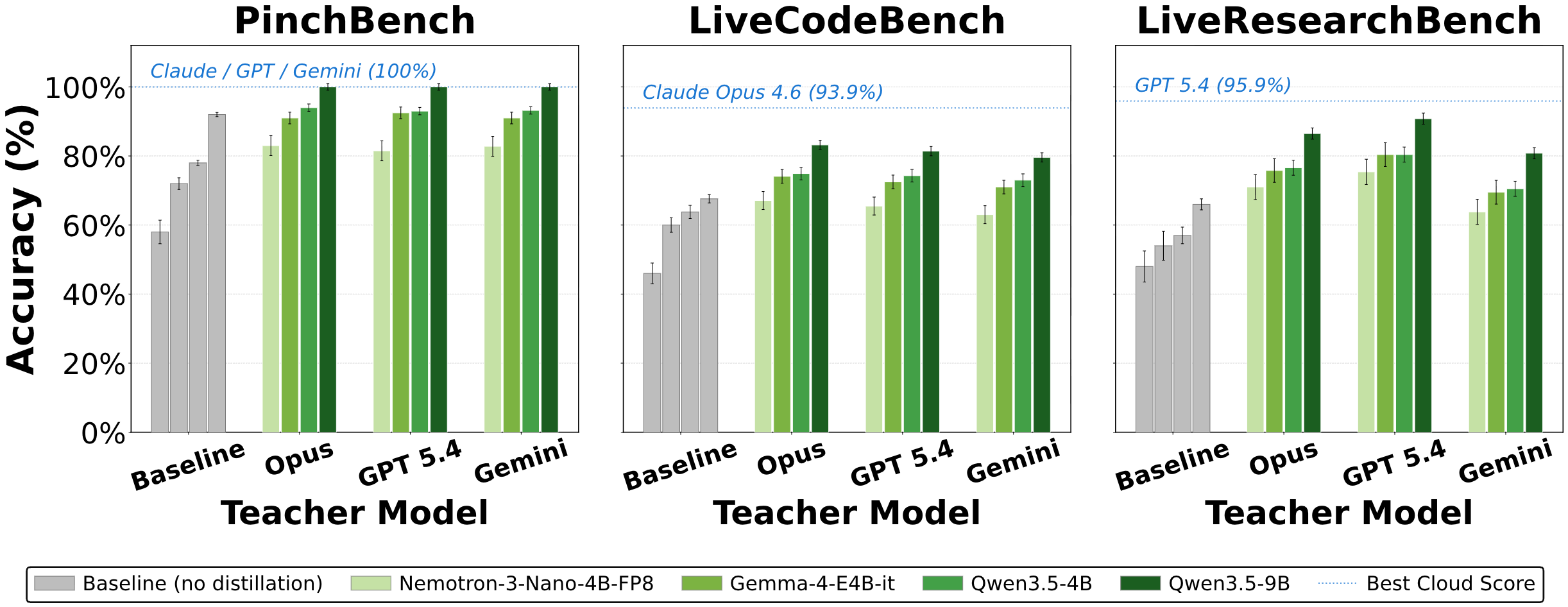}
    \caption{\textbf{\methodname improves local \ojSpecs.}
Every student--teacher pair improves over the unoptimized spec on all three benchmarks. The strongest search-optimized \texttt{Qwen3.5-9B} student reaches 100.0\% on \texttt{PinchBench}, 83.0\% on \texttt{LiveCodeBench}, and 91.0\% on \texttt{LiveResearchBench}.}
    \label{fig:spec_level_distillation} 
\end{figure}

\begin{figure}[b]
    \centering
    \includegraphics[width=\linewidth]{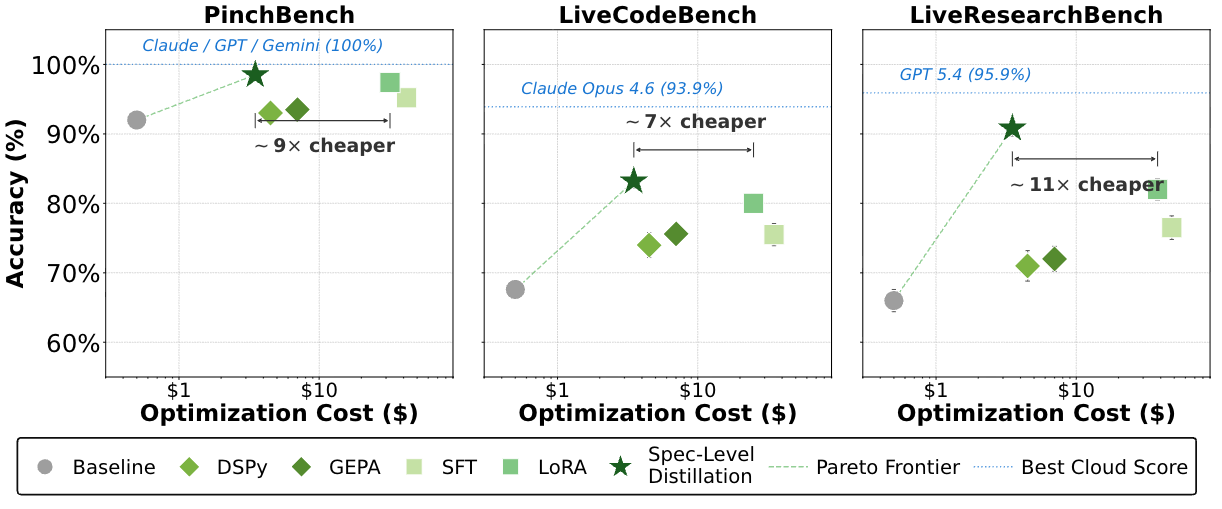}
    \caption{\textbf{Accuracy vs.\ optimization cost.}
\methodname reaches the best accuracy on all three main benchmarks.
LoRA is the strongest single-primitive baseline, but \methodname is 7.1--10.9$\times$ cheaper to optimize.
DSPy/SIMBA and prompt-only GEPA produce modest gains over the unoptimized local spec.}
    \label{fig:accuracy_vs_optimization_cost}
\end{figure}

\paragraph{\methodname improves OpenJarvis students on personal AI benchmarks.}
Figure~\ref{fig:spec_level_distillation} shows accuracy before and after search.
For \texttt{Qwen3.5-9B}, the best search-optimized spec reaches 100.0\% on \texttt{PinchBench}, 83.0\% on \texttt{LiveCodeBench}, and 91.0\% on \texttt{LiveResearchBench}.
Across the eight-benchmark suite, average gains per student model range from 13.1 to 31.5 pp (Appendix~\ref{app:full_distillation}, Table~\ref{tab:full_distillation}).
These gains reflect joint optimization over Intelligence, Agent, Tool, Memory, and Engine primitives rather than any single component.

\paragraph{\methodname improves accuracy at lower optimization cost.}
Figure~\ref{fig:accuracy_vs_optimization_cost} compares \methodname with prompt-only and weight-only baselines (DSPy/SIMBA, prompt-only GEPA, SFT, LoRA~\cite{khattab2024dspy, agrawal2026gepareflectivepromptevolution, hu2022lora}) under the same offline protocol.
The prompt-only baselines add 4.1--5.2~pp on average over the unoptimized local spec, and LoRA is the strongest weight-only baseline across all three panels.
\methodname adds 1.1--8.8~pp over LoRA and 5.0--18.8~pp over prompt-only GEPA at 7.1--10.9$\times$ lower optimization cost.
The cost advantage comes from the acceptance gate: rejected edits do not trigger full training runs, and accepted edits compound across sessions.

\begin{figure*}[t]
\centering
\includegraphics[width=\textwidth]{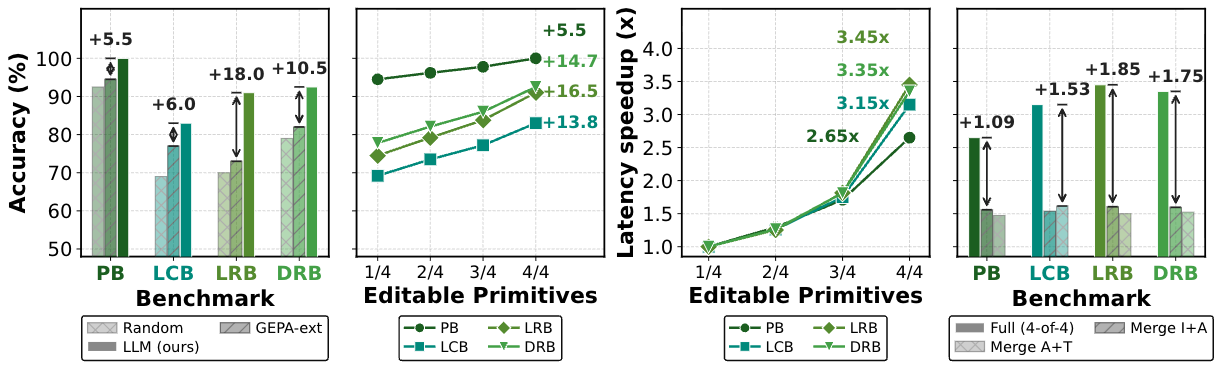}
\caption{\textbf{Proposer and move-space ablations for \methodname.}
Student: \texttt{Qwen3.5-9B}; teacher: \claudeOpus.
\textbf{Left:} at fixed four-primitive move space, the LLM proposer outperforms a template-random proposer and evolutionary spec search.
\textbf{Middle:} at fixed LLM proposer, expanding from one editable primitive to all four improves both accuracy and latency.
\textbf{Right:} merging primitive pairs reduces accuracy and speedup relative to the full four-primitive spec.
}
\label{fig:editable_set_ablation}
\end{figure*}

\paragraph{Both proposer and move space drive the gain.}
Figure~\ref{fig:editable_set_ablation} separates two axes: which proposer suggests edits, and how many primitives the proposer can edit.
With all four primitives editable, \methodname reaches 83.0--100.0\%, compared to 73.0--94.5\% for an evolutionary spec-search proposer at the same four-primitive move space (a gap of 5.5--18.0~pp, 10.0~pp on average) and 69.0--92.5\% for random sampling from the edit catalog (a gap of 7.5--21.0~pp, 14.0~pp on average).
The edit catalog and gate alone therefore do not explain the gain.
With the LLM teacher fixed as the proposer, expanding the editable set from one primitive to all four adds 5.5--16.5~pp in accuracy and 2.65--3.45$\times$ in latency speedup over the 1-of-4 baseline.
Merging primitive pairs that deployed systems treat as independent reverses these gains, dropping accuracy by 2.8--9.7~pp and latency speedup by 1.09--1.95$\times$ relative to the full 4-of-4 configuration.
Both axes matter: the LLM teacher needs the four-primitive spec to reach 83.0--100.0\%, and the four-primitive spec needs the LLM teacher to extract the full gain available at that move space.
Appendix~\ref{app:proposer_ablation} specifies the random and evolutionary spec-search four-primitive baselines.

\paragraph{What the optimizer learns.}
Accepted edits are distributed across all four editable primitives rather than concentrated on Intelligence weights: weight updates make up only 16--44\% of accepted edits across benchmarks, with the remainder spanning Engine, Agent, and Tool fields (Table~\ref{tab:edit_type_ablation}). 
The dominant primitive within that distribution tracks task structure: Intelligence edits dominate code (44\% on LiveCodeBench), Agent edits dominate customer-service and agentic tasks (41--45\% on PB, TauB, and TBTel), and Tool edits dominate tool-calling and research (39--47\% on TC15, GAIA, DRB, and LRB; Appendix~\ref{app:edit_type_ablation}, Figure~\ref{fig:edit_type_allocation}). 
At the per-failure level, the teacher maps each diagnosis to the corresponding primitive: retrieval failures receive mostly Tool edits, reasoning failures mostly Intelligence edits, control-flow failures mostly Agent edits, and efficiency-bounded failures mostly Engine edits (Figure~\ref{fig:cluster_edit_type}). 
This explains why no single primitive reaches the joint-search ceiling: the useful intervention depends on the failure mode, and many failures require coordinated changes across multiple primitives.

\section{Discussion and Conclusion}
\label{sec:discussion}
\label{sec:conclusion}

\sysname shows that local personal AI requires more than replacing a cloud model with an open-weight model.
The surrounding stack must be portable, measurable, and optimizable around the local model.
The spec abstraction provides the needed representation: it makes model substitution recoverable, exposes the local--cloud accuracy--efficiency frontier, and gives \methodname a target for transferring frontier-model guidance into an on-device system.
We close by clarifying what runs on-device, how frontier proposer models are used, and which robustness checks support the main claims.

\paragraph{Cloud at search time, local at inference time.}
The local--cloud division of labor in \methodname is deliberate: frontier cloud models are used at search time, where their capability advantage in reading traces and reasoning about coordinated edits matters most, and local hardware runs the resulting spec at inference time, where its latency, cost, and ownership properties matter most.
Capability flows from cloud to local through the optimized spec and stays there: the resulting local spec makes zero cloud calls at inference time, and undistilled local specs already match or beat cloud models on 4 of 8 benchmarks, so frontier proposer models extend local capability rather than being required for local systems to function.
When a frontier proposer is used at search time, only eligible traces are transmitted according to the protocol in Appendix~\ref{app:distillation_privacy}, and at 100 queries per day the amortized proposer cost falls below \$0.001 per query within six months (Appendix~\ref{app:teacher_amortization}).

\paragraph{Robustness and limitations.}
The headline gains from \methodname (Section~\ref{sec:experiments}) survive the four robustness perturbations we test: reward-weight variants, search-seed variance, random-restart gains, and proposer comparison all produce variation smaller than the headline effect (Appendix~\ref{app:distillation_robustness}).
Remaining limitations are statistical precision from 5 runs per configuration, possible judge bias from \texttt{GPT-5-mini}, and single-machine evaluation only (Appendix~\ref{app:limitations}).
Broader impacts and release safeguards are discussed in Appendix~\ref{app:broader_impacts}.

\section*{Acknowledgements}

We thank Kelly Buchanan, Francois Chaubard, Mayee Chen, Vivien Cheng, Catherine Deng, Neel Guha, Simon Guo, Braden Hancock, Junmiao Hu, Hangoo Kang, Andy Konwinski, Hermann Kumbong, Jacky Kwok, Adrian Lafuente Gamarra, Jerry Liu, Yuzhen Mao, Christopher Rytting, Tarun Suresh, Shayan Talaei, Roberto Torres, and Michael Zhang. We would also like to thank our collaborators at the Stanford Artificial Intelligence Laboratory (SAIL) and Stanford HAI.

We gratefully acknowledge support from federal sources: NIH under Nos.\ U54EB020405 (Mobilize) and R01GM124443 (SimTK); NSF under Nos.\ CCF2247015 (Hardware-Aware), CCF1763315 (Beyond Sparsity), CCF1563078 (Volume to Velocity), 1937301 (RTML), 1900638 (Open Federated Virtual Assistants), and 2028602 (PPoSS); DARPA under No.\ HR00112520038 (Fallingwater); US DEVCOM ARL under Nos.\ W911NF-23-2-0184 (Long-context) and W911NF-21-2-0251 (Interactive Human-AI Teaming); NSTXL under No.\ N00164-23-9-G057-01 (Energy-Efficient AI Hardware); and ONR under Nos.\ N000142312633 (Deep Signal Processing), N000142212426 (Misinformation Analysis), N000142012480 (Non-Euclidean Geometry), and N000142012275 (Data Programming).

We also gratefully acknowledge support from Stanford HAI under Nos.\ 247183, 215955, and 345077 (Evo); Google DeepMind, Google Research, and Google Cloud; member companies including NXP, Xilinx, LETI-CEA, Intel, IBM, Microsoft, NEC, Toshiba, TSMC, ARM, Hitachi, BASF, Accenture, Ericsson, Qualcomm, Analog Devices, Salesforce, and Total; the Laude Institute, Prime Intellect, Anthropic, and the HAI-GCP Cloud Credits for Research program; the Stanford Data Science Initiative (SDSI) and the Stanford Marlowe Computing Platform; the NSF Graduate Research Fellowship Program, the Knights-Hennessy Scholarship, the Stanford Graduate Fellowship, the JP Morgan AI/ML Fellowship, the Stanford EDGE Fellowship, and the GEM Fellowship; members of the Stanford DAWN project (Meta, Google, VMWare); and members of the Stanford SEAMS project (IBM, Felicis).

The U.S.\ Government is authorized to reproduce and distribute reprints for Governmental purposes notwithstanding any copyright notation thereon. Any opinions, findings, and conclusions or recommendations expressed in this material are those of the authors and do not necessarily reflect the views, policies, or endorsements, either expressed or implied, of NIH, ONR, DARPA, or the U.S.\ Government.

\begingroup
\sloppy
\bibliography{bibliography}
\bibliographystyle{plain}
\endgroup

\appendix

\section{Methods Details}
\label{app:methods}

\subsection{Primitive Implementation Details}
\label{app:primitives}

This section expands on the five primitives introduced in Section~\ref{sec:primitives}.

\paragraph{Intelligence: Model Catalog.}
The Intelligence catalog maintains metadata for each supported model: parameter count, context length, VRAM requirements, and compatible engines.
At runtime, the catalog merges static entries with models discovered from running backends (e.g., models pulled into \texttt{Ollama}), so the system always has a current view of what is locally available.
The user or the system can select the model best suited to the available hardware from this unified catalog.

\paragraph{Engine: Hardware Support.}
The hardware-detection module identifies the accelerator type and recommends the best Engine automatically.
Supported device classes include:
\begin{itemize}[leftmargin=1.5em, itemsep=2pt]
    \item \textbf{Phones}: iPhone, Google Pixel
    \item \textbf{Laptops}: MacBook Pro M4, AMD Ryzen AI
    \item \textbf{Workstations}: NVIDIA RTX 4090, AMD RX 7900 XTX
    \item \textbf{Dedicated AI hardware}: NVIDIA DGX Spark
\end{itemize}

\paragraph{Agents: Discrete and Continuous.}
\emph{Discrete Agents} perform actions only when called.
Available types include single-turn chat, multi-turn orchestration with function calling, ReAct~\cite{yao2023react} loops, and CodeAct-style~\cite{wang2024codeact} code execution.
\emph{Continuous Agents} run persistently, monitoring for events, maintaining state across sessions, and taking actions autonomously over long horizons.
Examples include scheduled morning digests that summarize email, calendar, and news; ongoing research tasks that monitor arXiv for relevant papers; and code review agents that watch GitHub repositories.

\paragraph{Tools \& Memory: Built-in Tools, Connectors, and Channels.}

\begin{table}[t]
\centering
\small
\begin{tabular}{llll}
\toprule
\textbf{Category} & \textbf{Tool} & \textbf{Runs} & \textbf{Cost / call} \\
\midrule
\multicolumn{4}{l}{\emph{Built-in Tools}} \\
Reasoning    & \texttt{think}                  & Local  & \$0 \\
Math         & \texttt{calculator}             & Local  & \$0 \\
Code         & \texttt{code\_interpreter}       & Local  & \$0 \\
Code         & \texttt{code\_interpreter\_docker} & Local  & \$0 \\
Code         & \texttt{repl}                   & Local  & \$0 \\
Search       & \texttt{web\_search}             & API    & \$0.005--0.01\textsuperscript{$\dagger$} \\
File I/O     & \texttt{file\_read}              & Local  & \$0 \\
HTTP         & \texttt{http\_request}           & Mixed  & Varies \\
Memory       & \texttt{retrieval}, \texttt{memory\_search}, \texttt{memory\_index} & Local & \$0 \\
Inference    & \texttt{llm}                    & Mixed  & \$0 local; API price cloud \\
Scheduler    & \texttt{schedule\_task}, \texttt{list/pause/resume/cancel} & Local & \$0 \\
Integration  & \texttt{mcp\_adapter}            & Mixed  & \$0 \\
\midrule
\multicolumn{4}{l}{\emph{Connectors (25+ data sources)}} \\
\multicolumn{4}{l}{Gmail, Google Calendar, Google Drive, Google Contacts, Google Tasks, Apple Health,} \\
\multicolumn{4}{l}{Apple Notes, Apple Music, Apple Contacts, iMessage, Notion, Obsidian, Slack,} \\
\multicolumn{4}{l}{Spotify, Strava, Oura, Outlook, Dropbox, GitHub, Hacker News, Weather, RSS, \ldots} \\
\midrule
\multicolumn{4}{l}{\emph{Channels (32+ messaging platforms)}} \\
\multicolumn{4}{l}{WhatsApp, Telegram, Discord, Slack, iMessage, SMS (Twilio), Signal, Teams,} \\
\multicolumn{4}{l}{Messenger, Email, IRC, Matrix, Mastodon, Reddit, Line, Viber, Webchat, \ldots} \\
\bottomrule
\end{tabular}

\caption{\textbf{Built-In Tools, Connectors, and Channels in \sysname}. \emph{Cost} indicates whether the tool incurs a per-call API fee; all other tools run locally at zero dollar cost. Connectors and channels are listed by count; full lists are in the documentation.}
\label{tab:tools}

\vspace{1pt}
{\footnotesize \textsuperscript{$\dagger$}Approximate per-query cost for Google Custom Search (\$5/1K queries), Tavily (\$0.005/query on paid plans), or Brave Search (\$0.009/query). Free tiers available with rate limits.}
\end{table}

Tools are the interfaces through which an Agent interacts with the outside world and the user's personal data.
Built-in Tools span seven categories: reasoning, math, code execution, web search, file I/O, memory, and inference delegation (Table~\ref{tab:tools}).
Web search is available via Google, Tavily, or Brave APIs; for on-device configurations, these tool API calls are the primary source of dollar cost, since local model inference has no marginal API fee.
An MCP~\cite{anthropic2024mcp} adapter allows any external MCP server to be used as a Tool.
Connectors pull data from 25+ sources (Gmail, Calendar, Apple Health, iMessage, Notion, Slack, and others); channels expose the Agent over 32+ messaging platforms (WhatsApp, Telegram, Discord, SMS, and others).
Memory is the persistent, searchable storage layer, with interchangeable backends: SQLite/FTS5 (default), FAISS~\cite{johnson2019faiss}, ColBERTv2~\cite{santhanam2022colbertv2}, BM25~\cite{robertson1994okapi}, and a hybrid using reciprocal rank fusion~\cite{cormack2009rrf}.

\paragraph{Learning: Training Data Sources.}
Intelligence-Based Learning can draw training data from three sources:
\begin{enumerate}[leftmargin=1.5em, itemsep=2pt]
    \item \textbf{Large-scale agentic datasets}: publicly available datasets such as GeneralThoughtArchive~\cite{turner2025generalthought} and ToolScale~\cite{su2025toolorchestra} provide diverse, verifiable queries spanning reasoning, tool use, and multi-turn interaction.
    \item \textbf{Synthetically generated traces}: the teacher model generates high-quality traces for specific query classes where the student underperforms.
    \item \textbf{Eligible user traces}: user-approved scrubbed interaction traces, annotated by the teacher or by ground-truth labels, provide signal grounded in the user's actual workload.
\end{enumerate}

Agent-Based Learning supports several optimization approaches:
\begin{itemize}[leftmargin=1.5em, itemsep=2pt]
    \item \textbf{DSPy-based prompt optimization}~\cite{khattab2024dspy}: compiles declarative signatures into optimized prompts via bootstrapped few-shot demonstrations.
    \item \textbf{Structured editing}: direct modification of system prompts, user prompt formatting, tool-use exemplars, and few-shot examples.
\end{itemize}

\paragraph{Spec format and TOML serialization.}
Section~\ref{sec:primitives} introduces the spec via the typed declaration in Figure~\ref{fig:spec_pseudocode}.
On disk, a spec is serialized as a TOML configuration that specifies:
\begin{itemize}[leftmargin=1.5em, itemsep=2pt]
    \item \textbf{Intelligence}: model identifier, temperature, top-$p$, max tokens, quantization format
    \item \textbf{Engine}: backend name, batch size, KV-cache settings
    \item \textbf{Agent}: agent type, system prompt, few-shot exemplars, turn limits
    \item \textbf{Tools \& Memory}: enabled Tools, Tool descriptions, Memory configuration, Connectors, Channels
    \item \textbf{Learning}: optimization approach, training data sources, reward weights
\end{itemize}
The primitives communicate through a thread-safe publish-subscribe EventBus, so adding a new component requires implementing only the relevant interface.
Figure~\ref{fig:spec_example} shows two example \ojSpecs that differ across all five primitives while sharing the same MCP tool layer and security configuration.

\begin{figure}[t]
\centering
\begin{minipage}[t]{0.44\linewidth}
\centering
\textbf{(a) Consumer deployment} \\[1pt]
\textit{Mac Mini M4, 24\,GB}
\vspace{1pt}
\begin{lstlisting}[language=TOML,basicstyle=\ttfamily\scriptsize,frame=single,framesep=2pt]
[intelligence]
default_model = "gemma4:4b-it"
quantization = "fp16"
max_tokens = 4096

[engine]
default = "ollama"

[agent]
default_agent = "simple"
max_turns = 10
tools = "think,calc,web_search"

[tools.storage]
default_backend = "sqlite"

[learning]
enabled = false
\end{lstlisting}
\end{minipage}%
\hspace{0.06\linewidth}%
\begin{minipage}[t]{0.44\linewidth}
\centering
\textbf{(b) Workstation deployment} \\[1pt]
\textit{NVIDIA H100, 80\,GB}
\vspace{1pt}
\begin{lstlisting}[language=TOML,basicstyle=\ttfamily\scriptsize,frame=single,framesep=2pt]
[intelligence]
default_model = "qwen3.5:122b"
quantization = "fp8"
max_tokens = 8192

[engine]
default = "vllm"

[agent]
default_agent = "native_openhands"
max_turns = 50
tools = "think,calc,code_interpreter,
         web_search,file_read,git_tool"

[tools.storage]
default_backend = "bm25"

[learning]
enabled = true
policy = "spec_distillation"
\end{lstlisting}
\end{minipage}
\caption{\textbf{Two \ojSpecs for the same personal AI system on different hardware.} The Consumer deployment serves Gemma4-4B via Ollama with a single-turn agent and learning disabled. The Workstation deployment serves Qwen3.5-122B (FP8) via vLLM with a multi-step coding agent, expanded tool set, and \methodname enabled. \texttt{[tools.mcp]}, \texttt{[security]}, and connectors (omitted) are identical across both.}
\label{fig:spec_example}
\end{figure}

\subsection{Privacy and Security Architecture}
\label{app:privacy}

This section expands on the privacy properties referenced in Section~\ref{sec:primitives}.

\paragraph{Privacy by architecture.}
Model inference, Engine execution, Agent state, telemetry, and eligible Learning updates run on-device.
Tools may cross the device boundary when connectors fetch from opted-in external services or the system calls an external API.
This separation provides privacy by construction, consistent with NIST AI RMF~\cite{nist2023airrmf}.
For ordinary inference, no traces, Memory contents, or training data leave the device.
\methodname is the exception: when a cloud teacher is enabled, only eligible scrubbed traces are transmitted during the bounded search phase (Appendix~\ref{app:distillation_privacy}).

\paragraph{Security at the Tools boundary.}
A security layer at the Tools boundary addresses the following risk categories from the OWASP Top~10 for LLM Applications~\cite{owasp2025llmtop10}:
\begin{itemize}[leftmargin=1.5em, itemsep=2pt]
    \item \textbf{Sensitive-data scanning}: outbound queries are scanned for PII, credentials, and other sensitive data before transmission.
    \item \textbf{Prompt-injection detection}: tool outputs are screened for adversarial instructions before being fed back to the model.
    \item \textbf{SSRF protection}: URL-based tool calls are validated against an allowlist to prevent server-side request forgery.
    \item \textbf{Sandboxed execution}: code interpreter tools run in isolated containers with restricted filesystem and network access.
\end{itemize}

\subsection{Cloud-Teacher Search Privacy}
\label{app:distillation_privacy}

\methodname transmits trace data to a teacher only during the search phase.
At inference time, the resulting local spec runs model inference and agent execution on-device.
Three controls bound data exposure during cloud-teacher search.

\paragraph{Scrubbing pipeline.}
Before any eligible trace is transmitted, it passes through the same sensitive-data scanner used at the Tools boundary (Appendix~\ref{app:privacy}).
The scanner redacts PII, credentials, and user-configured sensitive patterns, such as account numbers or medical identifiers.

\paragraph{Trace eligibility.}
Users specify which trace categories are eligible for cloud-teacher search.
By default, traces from Connectors marked sensitive are excluded.
Users can also mark specific conversations, sessions, or connectors as ineligible.

\paragraph{Local-teacher fallback.}
Users who require strict local-only operation can substitute a larger local model as the teacher.
This eliminates cloud transmission entirely, but we treat the resulting accuracy and convergence tradeoff as a deployment choice rather than the default setting evaluated in the main experiments.

\paragraph{Comparison to cloud personal AI.}
Frameworks like OpenClaw and Hermes Agent transmit inference queries to a cloud model by default.
\methodname confines cloud exposure to a bounded, optional search phase and applies a scrubbing layer before transmission.

\subsection{\methodname Details}
\label{app:distillation}

This section provides expanded examples for each phase of \methodname (Section~\ref{sec:distillation}).

\paragraph{Phase 1: Diagnose (Expanded).}
\label{app:diagnose}
The teacher LM ingests an eligible trace corpus, a dataset of structured JSON records drawn from benchmark traces, synthetic traces, or user-approved scrubbed traces.
Since traces are plain text, the teacher can read them directly, grep through fields, and run scripts over the JSON to identify patterns.
The teacher explores failure modes across query classes, compares its own outputs against the student's on held-out tasks, and produces failure clusters.
Each cluster is annotated with student vs.\ teacher success rates and a natural-language characterization of the skill gap (e.g., ``student fails on multi-hop questions requiring calendar lookups because it does not invoke the calendar tool'').

\paragraph{Phase 2: Plan (Expanded Edit Examples).}
The teacher proposes edits targeting specific primitives of the spec:

\begin{enumerate}[leftmargin=1.5em, itemsep=2pt]
    \item \textbf{Intelligence edits} modify the language model and its parameters.
    Examples include changing the model (e.g., switching from \texttt{Qwen3.5-4B} to \texttt{Qwen3.5-9B} for code queries), adjusting generation parameters (temperature, top-$p$, max tokens), changing the quantization format (e.g., Q4 to Q8), or triggering LoRA~\cite{hu2022lora} fine-tuning on teacher-generated SFT pairs or GRPO~\cite{shao2024deepseekmath} training with a composite reward (Equation~\ref{eq:reward}).
    \item \textbf{Engine edits} change the inference backend or its configuration.
    Examples include switching from \texttt{Ollama} to \texttt{vLLM} for higher throughput, adjusting batch size or KV-cache settings, or enabling a different quantization kernel.
    \item \textbf{Agent edits} modify the reasoning loop.
    Examples include replacing or refining the system prompt, editing few-shot exemplars, adjusting turn limits or verification steps, switching the Agent type (e.g., from single-turn to ReAct), or restructuring the tool-calling strategy.
    \item \textbf{Tools \& Memory edits} change the available Tools, their descriptions, and Memory configuration.
    Examples include adding or removing Tools from a given Agent, revising Tool descriptions to improve the model's tool selection, changing Memory configuration, and configuring a cloud model (\texttt{GPT~5.4}, \claudeOpus, \texttt{Gemini~3.1~Pro}) as an additional Tool when that setting is enabled.
\end{enumerate}

All edits are logged so that the teacher can examine its own intervention history across sessions and identify which changes helped vs.\ which did not.

\paragraph{Phase 3: Execute (Expanded Gate Description).}
The benchmark gate draws on three sources of signal:

\begin{enumerate}[leftmargin=1.5em, itemsep=2pt]
    \item \textbf{Synthetically annotated or eligible user traces}: the teacher annotates synthetic traces or user-approved scrubbed traces with ground-truth labels, providing signal grounded in the target workload.
    \item \textbf{Large-scale agentic datasets with known answers}: datasets such as GeneralThoughtArchive~\cite{turner2025generalthought} and ToolScale~\cite{su2025toolorchestra} provide diverse, verifiable queries spanning reasoning, tool use, and multi-turn interaction.
    \item \textbf{Standard benchmark splits}: train/test splits of established benchmarks such as MMLU-Pro~\cite{wang2024mmlupro}, GAIA~\cite{mialon2024gaia}, and $\tau$-bench~\cite{yao2024taubench} provide calibrated difficulty levels and reproducible evaluation.
\end{enumerate}

An edit is accepted if the updated spec improves on the target failure cluster without excessive regression on other clusters.
Rejected edits are logged with their gate results.

\section{Experimental Details}
\label{app:experiments}

\subsection{Benchmark Descriptions}
\label{app:benchmarks}

\begin{table}[ht]
\centering
\caption{Benchmark suite spanning eight personal AI workload categories (508 tasks total).}
\label{tab:benchmarks}
\small
\begin{tabular}{@{}llcl@{}}
\toprule
\textbf{Benchmark} & \textbf{Category} & \textbf{Tasks} & \textbf{Scoring} \\
\midrule
\texttt{ToolCall-15} & Tool calling & 15 & Automated \\
\texttt{PinchBench} & Agent tasks & 23 & Auto + LLM judge \\
\texttt{LiveCodeBench} (v6) & Coding & 100 & Automated \\
$\tau$-Bench V2 & Customer service & 100 & DB-state match \\
$\tau^2$-Bench Telecom & Customer service & 40 & DB-state match \\
\texttt{GAIA} & General assistant & 50 & Exact match \\
\texttt{LiveResearchBench} (v4) & Deep research & 100 & Checklist + LLM \\
\texttt{DeepResearchBench} & Deep research & 80 & RACE + FACT \\
\bottomrule
\end{tabular}
\end{table}

\begin{itemize}[leftmargin=*]

\item \texttt{ToolCall-15} (TC15)~\cite{toolcall15} evaluates single-turn tool-calling accuracy across 15 fixed scenarios organized into 5 categories.
Each scenario provides a mocked tool environment with deterministic expected behavior; scoring is fully automated (pass, partial, or fail).
We use all 15 scenarios.

\item \texttt{PinchBench} (PB)~\cite{pinchbench2025} measures end-to-end agent task completion across 23 tasks spanning calendar management, email handling, research, coding, and multi-step workflows.
Tasks are graded via a combination of automated checks and LLM-judge rubrics.
Originally developed as the canonical benchmark for the \texttt{OpenClaw} agent ecosystem~\cite{steinberger2025openclaw}, \texttt{PinchBench} tests practical agent competence rather than isolated model capabilities.

\item \texttt{LiveCodeBench} (LCB)~\cite{jain2024livecodebench} provides contamination-free competitive programming evaluation by continuously collecting problems from LeetCode, AtCoder, and CodeForces.
We use \texttt{release\_v6} (problems through April 2025; 1{,}055 total) and evaluate on a 100-problem subset sampled from problems released after January 2025 to minimize contamination risk.
Scoring is via automated test-case execution.

\item $\tau$-Bench~V2 (TauB)~\cite{yao2024taubench} simulates multi-turn customer service conversations between a simulated user and an agent equipped with domain-specific API tools and policy guidelines.
We evaluate on 100 tasks across the airline and retail domains, using database-state comparison for faithful automated evaluation.

\item $\tau^2$-Bench~Telecom (TBTel)~\cite{barres2025tau2bench} extends $\tau$-Bench to a dual-control telecom domain where both agent and user modify a shared environment.
We evaluate on 40 tasks.
This domain is particularly challenging because the agent must guide users through technical troubleshooting, requiring coordination beyond standard single-control customer service.

\item \texttt{GAIA}~\cite{mialon2024gaia} tests general AI assistant capabilities with 50 questions requiring web search, multi-step reasoning, and tool use.
Tasks range from simple factual lookups to complex multi-hop queries demanding synthesis across sources.
Evaluation uses exact-match scoring with a 2-hour timeout per task.

\item \texttt{LiveResearchBench} (LRB)~\cite{wang2025liveresearchbench} evaluates deep research capabilities with 100 expert-curated tasks spanning daily life, enterprise, and academia, each requiring extensive real-time web search and multi-source synthesis.
Built with over 1{,}500 hours of human labor, tasks are designed to be user-centric, dynamic, and unambiguous.
We evaluate on 100 tasks using the DeepEval checklist-based scoring protocol.
We use the v4 release (December 2025).

\item \texttt{DeepResearchBench} (DRB)~\cite{du2025deepresearchbenchcomprehensivebenchmark} evaluates deep research agents on 100 PhD-level tasks across 22 domains, with each task crafted by domain experts.
The task distribution reflects real-world research demand, derived from an analysis of over 96{,}000 user queries.
We evaluate on an 80-task subset.
Evaluation uses two complementary frameworks: RACE (Reference-based Adaptive Criteria-driven Evaluation), which scores generated reports against a reference across four dimensions (comprehensiveness, insight, instruction-following, and readability) using dynamically weighted, task-specific criteria; and FACT (Factual Abundance and Citation Trustworthiness), which measures citation accuracy and the number of verifiably supported claims.

\end{itemize}

\subsection{Model Selection Rationale}
\label{app:models}

\begin{table}[ht]
\centering
\caption{Models evaluated. \emph{Type} indicates architecture (Dense, MoE, or Hybrid Mamba-2). \emph{Active} denotes parameters per forward pass. \emph{Quant} indicates quantization format. \emph{Engine} is the inference backend within \sysname. \emph{License} is listed for reproducibility.}
\label{tab:models}
\small
\setlength{\tabcolsep}{3pt}
\begin{tabular}{@{}llcclc@{}}
\toprule
\textbf{Model} & \textbf{Type} & \textbf{Active} & \textbf{Quant} & \textbf{Engine} & \textbf{License} \\
\midrule
\multicolumn{6}{@{}l}{\emph{Local models: Qwen3.5 family}~\cite{qwen3.5}} \\
\texttt{Qwen3.5-4B} & Dense & 4B & FP16 & \texttt{Ollama} & Apache 2.0 \\
\texttt{Qwen3.5-9B} & Dense & 9B & FP16 & \texttt{vLLM} / \texttt{Ollama} & Apache 2.0 \\
\texttt{Qwen3.5-27B} & Dense & 27B & FP8 & \texttt{vLLM} & Apache 2.0 \\
\texttt{Qwen3.5-35B} & MoE & ${\sim}$3B & FP16 & \texttt{vLLM} / \texttt{Ollama} & Apache 2.0 \\
\texttt{Qwen3.5-122B} & MoE & ${\sim}$10B & FP8 & \texttt{vLLM} & Apache 2.0 \\
\midrule
\multicolumn{6}{@{}l}{\emph{Local models: Nemotron family}~\cite{nemotron_super_2025}} \\
\texttt{Nemotron-Nano-4B} & Hybrid & 4B & FP8 & \texttt{vLLM} & Nemotron Open \\
\texttt{Nemotron-Super-120B} & Hybrid MoE & ${\sim}$12B & FP8 & \texttt{vLLM} & Nemotron Open \\
\midrule
\multicolumn{6}{@{}l}{\emph{Local models: Gemma4 family}~\cite{gemma4_2025}} \\
\texttt{Gemma4-E4B} & Dense (PLE) & ${\sim}$4B & FP16 & \texttt{Ollama} & Apache 2.0 \\
\texttt{Gemma4-26B} & MoE & ${\sim}$4B & FP16 & \texttt{vLLM} & Apache 2.0 \\
\midrule
\multicolumn{6}{@{}l}{\emph{Local models: Granite family}~\cite{granite2025}} \\
\texttt{Granite~3.3~8B} & Dense & 8B & FP16 & \texttt{Ollama} & Apache 2.0 \\
\texttt{Granite~4.0~H-Small} & Hybrid MoE & ${\sim}$9B & FP16 & \texttt{vLLM} & Apache 2.0 \\
\midrule
\multicolumn{6}{@{}l}{\emph{Cloud baselines}} \\
\claudeOpus~\cite{claude_opus_46} & Undisc. & Undisc. & Undisc. & Cloud API & Proprietary \\
\texttt{GPT~5.4}~\cite{gpt54_2026} & Undisc. & Undisc. & Undisc. & Cloud API & Proprietary \\
\texttt{Gemini~3.1~Pro}~\cite{gemini31pro_2026} & Undisc. & Undisc. & Undisc. & Cloud API & Proprietary \\
\bottomrule
\end{tabular}

\vspace{2pt}
{\footnotesize \emph{Hybrid} = Mamba-2 + Transformer architecture. \emph{PLE} = Per-Layer Embeddings (Gemma4 edge models). \emph{Nemotron Open} = NVIDIA Nemotron Open Model License (commercial use permitted). For proprietary cloud APIs, Type, Active, and Quant are undisclosed. All Apache 2.0 and Nemotron Open models permit commercial use and derivative works.}
\end{table}

We select local models to cover the parameter-count spectrum relevant to consumer and workstation hardware, spanning four model families.
Priority is given to models demonstrating strong performance on at least one benchmark category.

\paragraph{Qwen3.5}~\cite{qwen3.5} provides the broadest coverage, with five variants from 4B to 122B.
\texttt{Qwen3.5-4B} represents compact laptop-class deployment and is used as a small student in the search experiments.
\texttt{Qwen3.5-9B} is a mid-range model that achieves the highest local results on \texttt{PinchBench} (96.8\%) and competitive results on $\tau$-Bench V2 (77.1\%).
\texttt{Qwen3.5-27B} (FP8), \texttt{Qwen3.5-35B} (${\sim}$3B active, MoE), and \texttt{Qwen3.5-122B} (${\sim}$10B active, MoE) progressively scale capacity.
\texttt{Qwen3.5-122B} achieves the best local average accuracy (80.3\%) and tops five of the eight benchmarks.

\paragraph{Nemotron}~\cite{nemotron_super_2025} contributes two models at opposite ends of the scale.
\texttt{Nemotron-Nano-4B} is a compact model for resource-constrained deployment.
\texttt{Nemotron-Super-120B} achieves the highest local \texttt{ToolCall-15} score (63.0\%), exceeding all three cloud baselines (max 53.3\%).

\paragraph{Gemma4}~\cite{gemma4_2025} contributes two models.
\texttt{Gemma4-26B} is included for its outlier \texttt{LiveCodeBench} performance (99.1\%), the highest score on that benchmark across all models including cloud, despite weak agentic scores.
\texttt{Gemma4-E4B} is a compact variant achieving 68.3\% on \texttt{LiveCodeBench}.

\paragraph{Granite}~\cite{granite2025} contributes two models from IBM's open model family.
\texttt{Granite~3.3~8B} and \texttt{Granite~4.0~H-Small} provide mid-range options for the main local-model sweep.

\paragraph{Cloud baselines.}
\claudeOpus~\cite{claude_opus_46} (Anthropic) achieves the highest average score (83.5\%) and leads on \texttt{GAIA} (62.0\%).
\texttt{GPT~5.4}~\cite{gpt54_2026} (OpenAI) achieves 95.9\% on \texttt{DeepResearchBench} and 100\% on both \texttt{PinchBench} and $\tau^2$-Bench~Telecom.
\texttt{Gemini~3.1~Pro}~\cite{gemini31pro_2026} (Google) achieves 100\% on \texttt{PinchBench} and 92.5\% on \texttt{LiveResearchBench}.

\paragraph{Per-benchmark leaders.}
No single local model leads on every benchmark.
\texttt{Qwen3.5-122B} tops 5 of 8 benchmarks (\texttt{PinchBench}, $\tau$-Bench~V2, $\tau^2$-Bench~Telecom, \texttt{DeepResearchBench}, \texttt{LiveResearchBench}) and the local average (80.3\%), but \texttt{Nemotron-Super-120B} leads on \texttt{ToolCall-15} (63.0\% vs.\ \texttt{Qwen3.5-122B} 60.0\%), \texttt{Gemma4-26B} leads on \texttt{LiveCodeBench} (99.1\% vs.\ 78.6\%), and \texttt{Qwen3.5-35B} leads on \texttt{GAIA} (55.7\% vs.\ 55.0\%).
This heterogeneity motivates the multi-model routing that \methodname produces in Section~\ref{sec:distillation_results}: a frontier teacher can diagnose per-benchmark failure modes and compose a spec that routes each query class to the local model best suited to it, rather than committing to a single model across the workload.

\subsection{Full Local-vs-Cloud Accuracy Table}
\label{app:full_accuracy_table}

Table~\ref{tab:main_table_results} reports per-benchmark accuracy for all evaluated local models and cloud baselines, supporting the headline portability and accuracy claims in Section~\ref{sec:portability}. The \emph{Best Local} row aggregates per-benchmark maxima across the primary local configurations, representing an oracle routing frontier.

\begin{table}[t]
  \centering
  \small
  \setlength{\tabcolsep}{3pt}
  \begin{tabular}{@{}lccccccccc@{}}
  \toprule
  \textbf{Model} & \textbf{TC15} & \textbf{PB} & \textbf{LCB} & \textbf{TauB} & \textbf{TBTel} & \textbf{GAIA} & \textbf{DRB} & \textbf{LRB} & \textbf{Avg} \\
  \midrule
  \multicolumn{10}{@{}l}{\emph{Cloud baselines}} \\
  \claudeOpus & 53.3 & \textbf{100.0} & 93.9 & 89.5 & 89.4 & \textbf{62.0} & 90.1 & 90.0 & \textbf{83.5} \\
  \texttt{GPT~5.4} & 46.6 & \textbf{100.0} & 70.0 & 89.2 & \textbf{100.0} & 33.3 & \textbf{95.9} & \textbf{96.2} & 78.9 \\
  \texttt{Gemini~3.1~Pro} & 53.3 & \textbf{100.0} & 80.0 & 90.8 & 85.0 & 51.3 & 85.2 & 92.5 & 79.8 \\
  \midrule
  \multicolumn{10}{@{}l}{\emph{Local models}} \\
  \texttt{Qwen3.5-9B} & 53.3 & 96.8 & 48.5 & 77.1 & 75.3 & 35.0 & 75.8 & 77.5 & 67.4 \\
  \texttt{Qwen3.5-27B} & 53.3 & \textbf{\underline{100.0}} & 51.0 & 88.4 & 84.9 & 50.4 & 77.6 & 82.5 & 73.5 \\
  \texttt{Qwen3.5-35B} & 60.0 & \textbf{\underline{100.0}} & 61.5 & 90.2 & 83.1 & \underline{55.7} & 79.4 & 86.9 & 77.1 \\
  \texttt{Qwen3.5-122B} & 60.0 & \textbf{\underline{100.0}} & 78.6 & \textbf{\underline{91.6}} & \underline{86.2} & 55.0 & \underline{80.5} & \underline{90.5} & \underline{80.3} \\ \midrule
  \texttt{Nemotron-Super-120B} & \textbf{\underline{63.0}} & 91.1 & 47.3 & 36.8 & 68.3 & 45.0 & 63.0 & 80.1 & 61.8 \\ \midrule
  \texttt{Gemma4-E4B} & 28.0 & 85.0 & 68.3 & 56.1 & 61.3 & 28.2 & 22.6 & 51.0 & 50.1 \\
  \texttt{Gemma4-26B} & 28.0 & 95.1 & \textbf{\underline{99.1}} & 91.3 & 78.5 & 52.1 & 72.1 & 84.2 & 75.1 \\ \midrule
  \texttt{Granite~3.3~8B} & 49.0 & 28.0 & 5.3 & 10.5 & 5.3 & 4.2 & 10.5 & 60.3 & 21.6 \\
  \texttt{Granite~4.0~H-Small} & 42.0 & 84.0 & 26.3 & 17.5 & 0.0 & 6.4 & 42.0 & 50.6 & 33.6 \\
  \midrule
  \multicolumn{10}{@{}l}{\emph{Best local (per-benchmark max across primary local configurations)}} \\
  \textbf{Best Local} & \underline{63.0} & \underline{100.0} & \underline{99.1} & \underline{91.6} & \underline{86.2} & \underline{55.7} & \underline{80.5} & \underline{90.5} & \underline{83.3} \\
  \bottomrule
  \end{tabular}
  \caption{\textbf{Full local-vs-cloud accuracy sweep.}
  \emph{Avg} is the unweighted mean across the 8 benchmarks.
  \textbf{Bold} = best score per benchmark across all models in this table; \underline{underline} = best local score.
  The best single local model, \texttt{Qwen3.5-122B}, reaches 80.3\% average accuracy, within 3.2~pp of the best cloud baseline, \claudeOpus at 83.5\%.
  The \emph{Best Local} row reports the per-benchmark maximum across the primary local configurations shown above, representing an oracle local routing frontier.
  Mean over 5 runs; full per-run statistics in Appendix~\ref{app:experiments}.}
  \label{tab:main_table_results}
\end{table}

\subsection{Hardware Specifications}
\label{app:hardware}

\begin{table}[ht]
\centering
\small
\begin{tabular}{@{}lllll@{}}
\toprule
\textbf{Vendor} & \textbf{Platform} & \textbf{Class} & \textbf{Memory} & \textbf{Energy API} \\
\midrule
NVIDIA & DGX Spark & AI Workstation & 128 GB LPDDR5X & \texttt{NVML} \\
NVIDIA & RTX 6000 Pro & Workstation GPU & 96 GB GDDR7 & \texttt{NVML} \\
AMD & Radeon RX 9070 XT & Consumer GPU & 16 GB GDDR6 & \texttt{ROCm SMI} \\
AMD & Ryzen AI Max+ 395 & Prosumer APU & 128 GB LPDDR5X & \texttt{ROCm SMI} \\
Intel & Arc Pro B70 & AI Workstation GPU & 32 GB GDDR6 & \texttt{xpu-smi} \\
Apple & Mac Mini M4 & Consumer & 16--32 GB & \texttt{powermetrics} \\
Apple & Mac Studio M4 Max & Workstation & 36--128 GB & \texttt{powermetrics} \\
\bottomrule
\end{tabular}
\caption{Hardware configurations. Seven platforms across four vendor families. Energy measurement uses vendor-specific APIs integrated into \sysname's telemetry module (Section~\ref{sec:evaluation}).}
\label{tab:hardware}
\end{table}

NVIDIA DGX~Spark\footnote{\url{https://www.nvidia.com/en-us/products/dgx/spark/}} is a compact AI workstation with a Grace-Blackwell Superchip delivering 1~PFLOPS of AI compute in a desktop form factor.
RTX~6000~Pro\footnote{\url{https://marketplace.nvidia.com/en-us/enterprise/laptops-workstations/nvidia-rtx-pro-6000-blackwell-workstation-edition/}} is a professional Blackwell workstation GPU with 96~GB GDDR7 memory and 24{,}064 CUDA cores.

AMD Radeon~RX~9070~XT\footnote{\url{https://www.amd.com/en/products/graphics/desktops/radeon/9000-series/amd-radeon-rx-9070xt.html}} is a consumer RDNA~4 GPU with 16~GB GDDR6 memory, representing the lower end of our hardware range.
Ryzen~AI~Max+~395\footnote{\url{https://www.amd.com/en/products/processors/laptop/ryzen/ai-300-series/amd-ryzen-ai-max-plus-395.html}} is a Zen~5 APU paired with Radeon~8060S integrated graphics and 128~GB unified LPDDR5X memory, evaluated in a Framework Desktop configuration.

Intel Arc~Pro~B70\footnote{\url{https://www.intel.com/content/www/us/en/products/sku/245797/intel-arc-pro-b70-graphics/specifications.html}} is a workstation-class Battlemage (Xe2-HPG) GPU with 32~GB GDDR6, 256 XMX engines, and 367 INT8 TOPS, targeting on-device AI inference at the consumer-discrete tier between consumer GPUs (16~GB) and high-end workstation cards (96~GB+).

Apple Mac~Mini~M4\footnote{\url{https://www.apple.com/mac-mini/}} is a consumer desktop with a 10-core CPU, 10-core GPU, and 16-core Neural Engine.
Mac~Studio~M4~Max\footnote{\url{https://www.apple.com/mac-studio/}} is a workstation with up to a 16-core CPU, 40-core GPU, and 128~GB unified memory.

\subsection{Evaluation Protocol}
\label{app:protocol}

Each (model, benchmark, hardware) configuration is run 5 times; we report mean accuracy and standard deviation.
For $\tau$-Bench, we additionally report pass$^k$ reliability metrics as defined in the original benchmark~\cite{yao2024taubench}.
Energy (joules), latency (seconds), power (watts), and dollar cost are recorded automatically by \sysname's instrumented inference wrapper and persisted to a local telemetry store (Section~\ref{sec:evaluation}).
All measurements are collected at batch size 1, consistent with interactive personal-AI workloads and matching the methodology of~\cite{saadfalcon2026intelligencewattmeasuringintelligence}.
Cloud energy is estimated from datacenter-level figures following the methodology of~\cite{saadfalcon2026intelligencewattmeasuringintelligence}.

\section{Discussion Details}
\label{app:discussion_details}

This appendix provides supporting evidence for the Discussion section claims.
Appendix~\ref{app:teacher_amortization} quantifies the amortized cost of teacher API usage during \methodname, supporting the Discussion point that cloud teachers are not a persistent dependency.
Appendix~\ref{app:hardware_profiling} profiles inference performance across the seven hardware platforms from Section~\ref{sec:setup}, supporting the Discussion point that on-device deployment scales from consumer to workstation hardware.
Appendix~\ref{app:full_distillation} extends the three-benchmark search results from Section~\ref{sec:distillation_results} to all eight benchmarks, supporting the Discussion point that the 13--32~pp improvement pattern generalizes across the full suite.
Appendix~\ref{app:edit_type_ablation} reports the edit-type distribution and single-primitive ablation that validates the optimization across primitives claim in Section~\ref{sec:distillation_results}.
Appendix~\ref{app:distillation_robustness} reports robustness checks on the 13--32~pp search gain across reward-weight perturbations and search seeds.

\subsection{Search over Proposer-Defined Neighborhoods}
\label{app:proposer_neighborhoods}

Classical hill climbing is usually analyzed over a fixed local neighborhood.
Each move changes a small part of the state, so greedy search can be trapped when every local neighbor is worse.
The spec search space is different because the neighborhood is proposer-defined.
A single LLM proposal can modify multiple typed slots at once.
For example, it can pair a tool-description rewrite with a prompt change, an Engine switch, and a generation-parameter update.
This does not make the space formally fully connected in the graph-theoretic sense, and we do not claim an optimality guarantee.
Instead, it changes the empirical failure mode.
The question is whether the proposer reliably suggests useful compound edits when improvements exist.
The proposer and editable-set ablations in Figure~\ref{fig:editable_set_ablation} show that both parts are needed in our setting.
Evolutionary spec search benefits from the four-primitive space but remains below the LLM-guided greedy proposer.
Restricting the LLM proposer to fewer editable primitives also leaves substantial accuracy on the table.

\subsection{Teacher Cost Amortization}
\label{app:teacher_amortization}

\begin{table}[ht]
\centering
\caption{Amortized teacher cost per query under \methodname. The one-time search cost (\$15.6 per benchmark) is amortized over the total number of inference queries during deployment. At even modest query volumes, the per-query teacher cost is negligible compared to cloud API pricing.}
\label{tab:teacher_amortization}
\small
\begin{tabular}{@{}lccccc@{}}
\toprule
\textbf{Deployment} & \textbf{Queries/day} & \textbf{Total queries} & \textbf{Search cost} & \textbf{Amortized/query} & \textbf{vs.\ Cloud API} \\
\midrule
1 week   & 100 & 700     & \$15.6 & \$0.0223  & 2.5$\times$ more expensive \\
1 month  & 100 & 3{,}000  & \$15.6 & \$0.0052  & 1.7$\times$ cheaper \\
6 months & 100 & 18{,}000 & \$15.6 & \$0.0009  & 10.4$\times$ cheaper \\
1 year   & 100 & 36{,}500 & \$15.6 & \$0.0004  & 21.1$\times$ cheaper \\
\midrule
1 week   & 200 & 1{,}400  & \$15.6 & \$0.0111  & 1.2$\times$ more expensive \\
1 month  & 200 & 6{,}000  & \$15.6 & \$0.0026  & 3.5$\times$ cheaper \\
6 months & 200 & 36{,}000 & \$15.6 & \$0.0004  & 20.8$\times$ cheaper \\
\bottomrule
\end{tabular}
\vspace{4pt}
\begin{minipage}{\linewidth}
{\footnotesize \textbf{Note}: Cloud API comparison uses \claudeOpus at an average cost of \$0.009 per query measured across our 8-benchmark suite (pricing per \href{https://platform.claude.com/docs/en/about-claude/pricing}{Anthropic's published rates}). Search cost of \$15.6 is the median across the three benchmarks in Section~\ref{sec:distillation_results}. Amortized cost per query equals search cost divided by total queries; ``cheaper'' ratio is \$0.009 divided by amortized cost.}
\end{minipage}
\end{table}

Table~\ref{tab:teacher_amortization} quantifies how the one-time teacher API cost of \methodname amortizes over deployment lifetime.
A single search session costs \$15.6 per benchmark in teacher API fees (median across the three benchmarks in Section~\ref{sec:distillation_results}), paid once to frontier providers (OpenAI, Anthropic, or Google) during the search phase.
After search, the resulting local-only spec makes no further teacher calls and runs local model inference on-device.
At 100 queries per day, the amortized teacher cost drops below \$0.001 per query within six months of deployment (\$0.0009) and below \$0.0005 within a year (\$0.0004).
At 200 queries per day, the amortization crosses \$0.0005 per query within six months.
In both regimes, the amortized per-query teacher cost becomes more than an order of magnitude cheaper than the \$0.009 per-query cost of running the equivalent workload against \claudeOpus directly.
For deployments shorter than roughly one month, search is more expensive than direct cloud API usage; the cost advantage materializes at longer deployment horizons.
For a one-year deployment at modest query volume (100/day), \methodname is 21$\times$ cheaper than equivalent cloud API usage; for higher-volume deployments (200/day over six months), the advantage exceeds 20$\times$.
These ratios assume the same teacher API is invoked for every query under the cloud baseline, which is the default behavior of frameworks like OpenClaw~\cite{steinberger2025openclaw} and Hermes Agent~\cite{nousresearch2025hermes}.

\subsection{Per-Platform Inference Profiling}
\label{app:hardware_profiling}

\begin{table}[h]
\centering
\scriptsize
\setlength{\tabcolsep}{1pt}
\begin{tabular}{@{}llccccccc@{}}
\toprule
\textbf{Platform} & \textbf{Model} & \textbf{Max Ctx} & \textbf{Prefill} & \textbf{Decode} & \textbf{Tok/s} & \textbf{Energy} & \textbf{Avg Acc.} & \textbf{HW Cost} \\
 & & & \textbf{(ms)} & \textbf{(ms/tok)} & \textbf{(out)} & \textbf{(J/query)} & \textbf{(\%)} & \textbf{(\$)} \\
\midrule
\multicolumn{9}{@{}l}{\emph{Consumer}} \\
Mac Mini M4 (24GB) & \texttt{Qwen3.5-9B} & 256K & 89{,}912 & 106 & 9.5 & 27{,}175 & 68.4 & \href{https://www.apple.com/shop/buy-mac/mac-mini}{\$999} \\
Mac Mini M4 (24GB) & \texttt{Gemma4-E4B} & 256K & 39{,}961 & 45.4 & 22.0 & 11{,}744 & 16.3 & \href{https://www.apple.com/shop/buy-mac/mac-mini}{\$999} \\
\midrule
Radeon RX 9070 XT & \texttt{Qwen3.5-9B} & 256K & 1{,}895 & 19.8 & 50.5 & 20{,}191 & 68.4 & \href{https://www.amd.com/en/products/graphics/desktops/radeon/9000-series/amd-radeon-rx-9070xt.html}{\$599} \\
Radeon RX 9070 XT & \texttt{Gemma4-E4B} & 256K & 842.4 & 8.51 & 118 & 8{,}681 & 16.3 & \href{https://www.amd.com/en/products/graphics/desktops/radeon/9000-series/amd-radeon-rx-9070xt.html}{\$599} \\
\midrule
\multicolumn{9}{@{}l}{\emph{AI-focused Discrete}} \\
Arc Pro B70 (32GB) & \texttt{Qwen3.5-9B} & 256K & 4{,}029 & 20.8 & 48.0 & 16{,}454 & 68.4 & \href{https://www.intel.com/content/www/us/en/products/sku/245797/intel-arc-pro-b70-graphics/specifications.html}{\$949} \\
Arc Pro B70 (32GB) & \texttt{Qwen3.5-27B} & 256K & 12{,}087 & 60.2 & 16.6 & 47{,}595 & 78.8 & \href{https://www.intel.com/content/www/us/en/products/sku/245797/intel-arc-pro-b70-graphics/specifications.html}{\$949} \\
Arc Pro B70 (32GB) & \texttt{Gemma4-26B} & 256K & 1{,}791 & 9.69 & 103 & 7{,}632 & 23.0 & \href{https://www.intel.com/content/www/us/en/products/sku/245797/intel-arc-pro-b70-graphics/specifications.html}{\$949} \\
\midrule
\multicolumn{9}{@{}l}{\emph{Prosumer / Workstation}} \\
Ryzen AI Max (128GB) & \texttt{Qwen3.5-27B} & 256K & 37{,}489 & 143 & 7.0 & 59{,}820 & 78.8 & \href{https://frame.work/desktop}{\$1{,}999} \\
Ryzen AI Max (128GB) & \texttt{Nemotron-Super-120B} & 256K & 16{,}662 & 60.0 & 16.7 & 25{,}187 & 52.1 & \href{https://frame.work/desktop}{\$1{,}999} \\
Ryzen AI Max (128GB) & \texttt{Gemma4-26B} & 256K & 5{,}554 & 23.0 & 43.5 & 9{,}582 & 23.0 & \href{https://frame.work/desktop}{\$1{,}999} \\
\midrule
Mac Studio M4 Max (128GB) & \texttt{Qwen3.5-122B} & 256K & 6{,}206 & 22.1 & 45.2 & 10{,}834 & 78.8 & \href{https://www.apple.com/shop/buy-mac/mac-studio}{\$3{,}499} \\
Mac Studio M4 Max (128GB) & \texttt{Nemotron-Super-120B} & 256K & 7{,}447 & 25.0 & 40.0 & 12{,}308 & 52.1 & \href{https://www.apple.com/shop/buy-mac/mac-studio}{\$3{,}499} \\
Mac Studio M4 Max (128GB) & \texttt{Gemma4-26B} & 256K & 2{,}482 & 9.59 & 104 & 4{,}680 & 23.0 & \href{https://www.apple.com/shop/buy-mac/mac-studio}{\$3{,}499} \\
\midrule
RTX 6000 Pro & \texttt{Qwen3.5-35B-A3B} & 256K & 245.8 & 2.59 & 386 & 5{,}210 & 78.8 & \href{https://marketplace.nvidia.com/en-us/enterprise/laptops-workstations/nvidia-rtx-pro-6000-blackwell-workstation-edition/}{\$8{,}900} \\
RTX 6000 Pro & \texttt{Nemotron-Nano-Omni-30B-A3B} & 256K & 245.8 & 2.29 & 436 & 4{,}623 & 52.1 & \href{https://marketplace.nvidia.com/en-us/enterprise/laptops-workstations/nvidia-rtx-pro-6000-blackwell-workstation-edition/}{\$8{,}900} \\
RTX 6000 Pro & \texttt{Gemma4-26B} & 256K & 327.7 & 3.29 & 304 & 6{,}621 & 23.0 & \href{https://marketplace.nvidia.com/en-us/enterprise/laptops-workstations/nvidia-rtx-pro-6000-blackwell-workstation-edition/}{\$8{,}900} \\
\midrule
\multicolumn{9}{@{}l}{\emph{AI Workstation}} \\
DGX Spark & \texttt{Qwen3.5-122B} & 256K & 3{,}277 & 49.7 & 20.1 & 23{,}169 & 78.8 & \href{https://marketplace.nvidia.com/en-us/enterprise/personal-ai-supercomputers/dgx-spark/}{\$4{,}699} \\
DGX Spark & \texttt{Nemotron-Super-120B} & 256K & 3{,}932 & 56.3 & 17.8 & 26{,}246 & 52.1 & \href{https://marketplace.nvidia.com/en-us/enterprise/personal-ai-supercomputers/dgx-spark/}{\$4{,}699} \\
DGX Spark & \texttt{Gemma4-26B} & 256K & 1{,}311 & 21.6 & 46.3 & 10{,}047 & 23.0 & \href{https://marketplace.nvidia.com/en-us/enterprise/personal-ai-supercomputers/dgx-spark/}{\$4{,}699} \\
\bottomrule
\end{tabular}
\caption{\textbf{Inference performance by hardware platform for representative local models} (batch size 1; 32K input plus 4K output; see Appendix~\ref{app:protocol}).
Each platform runs the three largest models that fit cleanly at FP8 quantization, spanning multiple model families where capacity allows.
Prefill and decode times are medians across all benchmark queries.
\emph{Hardware cost} is approximate retail price as of April 2026.}
\label{tab:hardware_profiling_b1}
\end{table}

Table~\ref{tab:hardware_profiling_b1} profiles inference performance for representative local models on each of the seven hardware platforms from Section~\ref{sec:setup}, at batch size 1 (interactive personal-AI workload, matching the measurement protocol in Appendix~\ref{app:protocol}).
Each platform is evaluated on the three largest models that fit cleanly at FP8 quantization, spanning the Qwen3.5, Nemotron, Gemma4, and Granite families.
Consumer platforms (Mac~Mini~M4 at \$999, Radeon~RX~9070~XT at \$599) run mid-range models (\texttt{Qwen3.5-9B}, \texttt{Granite~3.3~8B}, \texttt{Gemma4-E4B}) that each fit within 16--24~GB of memory.
Prosumer and workstation platforms (Ryzen~AI~Max at \$1{,}999, Mac~Studio~M4~Max at \$3{,}499, RTX~6000~Pro at \$8{,}900) support the largest local models (\texttt{Qwen3.5-122B}, \texttt{Nemotron-Super-120B}, \texttt{Gemma4-26B}) at full precision.
The DGX~Spark AI workstation (\$4{,}699) provides the highest per-query throughput at the largest model sizes.
Across platforms, the spec abstraction enables the same Agent and Tool configuration to be retargeted to different hardware tiers without rewriting prompts or agent logic: only the \texttt{[intelligence]} and \texttt{[engine]} slots change (Figure~\ref{fig:spec_example}).
The accuracy column reports the served model configuration used for the corresponding hardware profile. Hardware affects latency, throughput, and energy; accuracy differences arise from the selected model, quantization, runtime, and evaluation configuration rather than the hardware alone.

\subsection{Full Search Results Across All Benchmarks}
\label{app:full_distillation}

\begin{table}[ht]
\centering
\caption{\methodname accuracy (\%) across all 8 benchmarks with the best teacher per benchmark.
\emph{Baseline} is the unoptimized student accuracy; \emph{Optimized} is the best result across 3 teachers; \emph{$\Delta$} is the gain.
Section~\ref{sec:distillation_results} reports detailed per-teacher results for PB, LCB, and LRB; this table extends those results to the full benchmark suite.}
\label{tab:full_distillation}
\small
\setlength{\tabcolsep}{3pt}
\begin{tabular}{@{}l ccc ccc ccc ccc@{}}
\toprule
& \multicolumn{3}{c}{\texttt{Nemotron-Nano-4B}} & \multicolumn{3}{c}{\texttt{Gemma4-E4B}} & \multicolumn{3}{c}{\texttt{Qwen3.5-4B}} & \multicolumn{3}{c}{\texttt{Qwen3.5-9B}} \\
\cmidrule(lr){2-4} \cmidrule(lr){5-7} \cmidrule(lr){8-10} \cmidrule(lr){11-13}
\textbf{Bench.} & Base & Opt. & $\Delta$ & Base & Opt. & $\Delta$ & Base & Opt. & $\Delta$ & Base & Opt. & $\Delta$ \\
\midrule
TC15   & 40.0 & 48.8 & +8.8  & 28.0 & 35.2 & +7.2  & 46.7 & 54.1 & +7.4  & 53.3 & 62.5  & +9.2  \\
PB     & 8.7  & 83.0 & +74.3 & 85.0 & 96.5 & +11.5 & 91.3 & 94.0 & +2.7  & 96.8 & 100.0 & +3.2  \\
LCB    & 10.0 & 67.0 & +57.0 & 68.3 & 81.5 & +13.2 & 10.0 & 75.0 & +65.0 & 48.5 & 83.0  & +34.5 \\
TauB   & 11.1 & 25.6 & +14.5 & 56.1 & 72.4 & +16.3 & 44.4 & 56.7 & +12.3 & 77.1 & 91.8  & +14.7 \\
TBTel  & 0.0  & 11.4 & +11.4 & 61.3 & 75.0 & +13.7 & 0.0  & 9.4  & +9.4  & 75.3 & 90.5  & +15.2 \\
GAIA   & 8.0  & 16.1 & +8.1  & 28.2 & 38.6 & +10.4 & 22.0 & 28.7 & +6.7  & 35.0 & 47.3  & +12.3 \\
DRB    & 0.0  & 15.6 & +15.6 & 22.6 & 38.2 & +15.6 & 50.0 & 63.1 & +13.1 & 75.8 & 92.5  & +16.7 \\
LRB    & 12.5 & 75.0 & +62.5 & 51.0 & 68.5 & +17.5 & 13.8 & 80.0 & +66.2 & 77.5 & 91.0  & +13.5 \\
\midrule
\textbf{Avg}   & 11.3 & 42.8 & +31.5 & 50.1 & 63.2 & +13.1 & 34.8 & 57.6 & +22.9 & 67.4 & 82.3 & +14.9 \\
\bottomrule
\end{tabular}
\vspace{4pt}
\begin{minipage}{\textwidth}
{\footnotesize Baselines for \texttt{Qwen3.5-9B} and \texttt{Gemma4-E4B} are reproduced from Table~\ref{tab:main_table_results}. Baselines for \texttt{Nemotron-Nano-4B} and \texttt{Qwen3.5-4B} are from the extended evaluation in Appendix~\ref{app:protocol} (these models are not included in the main table). PB, LCB, and LRB optimized values are from Section~\ref{sec:distillation_results} (best teacher per benchmark). All averages computed across the 8 benchmarks. All values use 5 independent runs; means reported.}
\end{minipage}
\end{table}

Table~\ref{tab:full_distillation} extends the three-benchmark search evaluation from Section~\ref{sec:distillation_results} (PB, LCB, LRB) to all eight benchmarks.
For each (student, benchmark) pair, we report the unoptimized baseline accuracy, the best search-optimized accuracy across the three teachers (\claudeOpus, \texttt{GPT~5.4}, \texttt{Gemini~3.1~Pro}), and the gain.
The 13--32~pp improvement pattern from Section~\ref{sec:distillation_results} holds across the full benchmark suite when computed on student-level averages: \texttt{Nemotron-Nano-4B} improves by +31.5~pp, \texttt{Qwen3.5-4B} by +22.9~pp, \texttt{Gemma4-E4B} by +13.1~pp, and \texttt{Qwen3.5-9B} by +14.9~pp on average.
The gains demonstrate that \methodname can convert phone- and laptop-scale models into viable personal AI substrates.
On \texttt{GAIA}, where the unoptimized cloud--local gap is 27.0~pp (Section~\ref{sec:portability}), search narrows the \texttt{Qwen3.5-9B} gap to cloud (\claudeOpus at 62.0\%) from 27.0~pp (baseline 35.0\%) to 14.7~pp (optimized 47.3\%).
On \texttt{DeepResearchBench}, where the unoptimized gap is 20.1~pp, search narrows the \texttt{Qwen3.5-9B} gap to cloud (\texttt{GPT~5.4} at 95.9\%) from 20.1~pp (baseline 75.8\%) to 3.4~pp (optimized 92.5\%).
These results confirm that \methodname is effective beyond the three benchmarks used as the primary demonstration; the method generalizes to tool-calling, customer service, and reasoning tasks as well as agent, code, and research tasks.

\paragraph{Different teachers specialize.}
No single teacher is uniformly best across benchmarks.
\texttt{GPT~5.4} is the strongest teacher on \texttt{LiveResearchBench}, producing a 91\% optimized \texttt{Qwen3.5-9B} student versus 86\% with \claudeOpus and 81\% with \texttt{Gemini~3.1~Pro}, consistent with GPT~5.4's lead on the cloud-only \texttt{LiveResearchBench} result (96.2\%).
\claudeOpus is the strongest teacher on \texttt{LiveCodeBench}, producing an 83\% optimized student versus 81\% with \texttt{GPT~5.4} and 80\% with \texttt{Gemini~3.1~Pro}, consistent with Claude's lead on the cloud-only \texttt{LiveCodeBench} result (93.9\%).
On \texttt{PinchBench}, all three teachers produce equivalent 100\% optimized students because all three cloud models themselves saturate at 100\%.
Because the spec treats teacher selection as a configurable Learning parameter, a practical deployment can route each query class to the teacher best suited to it during search, without committing to a single teacher across the workload.

\subsection{Edit-Type Distribution and Ablation}
\label{app:edit_type_ablation}

\begin{table}[ht]
\centering
\caption{Distribution of accepted edits by primitive type during \methodname for \texttt{Qwen3.5-9B} with \claudeOpus as teacher across all eight benchmarks. \emph{Fraction} is the share of total accepted edits targeting that primitive. \emph{Acc.} reports solo accuracy for primitive rows and full four-primitive accuracy for the Full search row; the gap to full \methodname quantifies the contribution of cross-component optimization. Engine edits modify the inference runtime (backend, batch size, KV-cache) and affect efficiency (latency, energy, throughput) but not accuracy; we mark Acc. as N/A and report only the fraction of accepted edits.}
\label{tab:edit_type_ablation}
\scriptsize
\setlength{\tabcolsep}{2.5pt}
\begin{tabular}{@{}l cc cc cc cc cc cc cc cc@{}}
\toprule
& \multicolumn{2}{c}{\texttt{TC15}} & \multicolumn{2}{c}{\texttt{PB}} & \multicolumn{2}{c}{\texttt{LCB}} & \multicolumn{2}{c}{\texttt{TauB}} & \multicolumn{2}{c}{\texttt{TBTel}} & \multicolumn{2}{c}{\texttt{GAIA}} & \multicolumn{2}{c}{\texttt{DRB}} & \multicolumn{2}{c}{\texttt{LRB}} \\
\cmidrule(lr){2-3} \cmidrule(lr){4-5} \cmidrule(lr){6-7} \cmidrule(lr){8-9} \cmidrule(lr){10-11} \cmidrule(lr){12-13} \cmidrule(lr){14-15} \cmidrule(lr){16-17}
\textbf{Edit type} & Frac. & Acc. & Frac. & Acc. & Frac. & Acc. & Frac. & Acc. & Frac. & Acc. & Frac. & Acc. & Frac. & Acc. & Frac. & Acc. \\
\midrule
Intelligence  & 16\% & 54.5 & 18\% & 96.9 & 44\% & 69.2 & 18\% & 79.5 & 17\% & 76.0 & 19\% & 35.5 & 17\% & 76.5 & 22\% & 78.0 \\
Agent         & 24\% & 55.6 & 41\% & 94.5 & 28\% & 66.3 & 43\% & 83.0 & 45\% & 80.0 & 28\% & 37.0 & 23\% & 77.5 & 25\% & 79.0 \\
Tool          & 47\% & 56.5 & 29\% & 97.0 & 14\% & 64.6 & 27\% & 81.7 & 26\% & 78.2 & 41\% & 39.0 & 47\% & 77.8 & 39\% & 74.5 \\
Engine        & 13\% & N/A  & 12\% & N/A  & 14\% & N/A  & 12\% & N/A  & 12\% & N/A  & 12\% & N/A  & 13\% & N/A  & 14\% & N/A  \\
\midrule
Full search & N/A & 62.5 & N/A & 100.0 & N/A & 83.0 & N/A & 91.8 & N/A & 90.5 & N/A & 47.3 & N/A & 92.5 & N/A & 91.0 \\
\bottomrule
\end{tabular}
\vspace{2pt}
\begin{minipage}{\linewidth}
{\footnotesize \emph{Acc.} is solo accuracy for primitive rows, obtained by restricting the teacher to one edit type, and full four-primitive accuracy for the Full search row. The gap between the best solo row and full search quantifies the value of cross-component optimization. Engine edits affect efficiency rather than accuracy and are reported as N/A; the Fraction column still reflects accepted Engine edits during full search. All accuracy values are \%.}
\end{minipage}
\end{table}

Table~\ref{tab:edit_type_ablation} provides the full data supporting the optimization across primitives claim in Section~\ref{sec:distillation_results}.
For \texttt{Qwen3.5-9B} optimized with \claudeOpus as teacher, we report two quantities per primitive: the fraction of all accepted edits during full search that targeted that primitive (\emph{Fraction}) and the accuracy achieved when the teacher is restricted to proposing only edits of that primitive type (\emph{Acc.}, with all other edit types disabled).
The gap between the highest solo accuracy and the full-search accuracy quantifies the contribution of search across primitives.

Two patterns validate optimizing across primitives.
First, no single primitive reaches the full-search ceiling on any benchmark: the best solo configuration is 5.5~pp below full search on \texttt{PinchBench} (Agent solo 94.5\% vs.\ 100\%), 13.8~pp below on \texttt{LiveCodeBench} (Intelligence solo 69.2\% vs.\ 83\%), and 16.5~pp below on \texttt{LiveResearchBench} (Tool solo 74.5\% vs.\ 91\%).
Joint search across primitives therefore adds 5.5--16.5~pp on top of the best single-primitive configuration.
Second, the dominant primitive varies by task type in a way consistent with task structure: agent tasks (PinchBench) favor Agent and Tool edits, code tasks (LiveCodeBench) favor Intelligence edits (consistent with code generation being primarily weight-dependent), and research tasks (LiveResearchBench) favor Tool edits (consistent with deep research depending on retrieval and tool selection).

The fraction row confirms that the teacher's allocation behavior during full search tracks these ablation results: on \texttt{LiveCodeBench}, Intelligence edits account for 44\% of all accepted edits (the highest share), matching Intelligence's position as the best-performing solo primitive; on \texttt{LiveResearchBench}, Tool edits account for 39\% (the highest share), matching Tool's position as the best-performing solo primitive.
The teacher allocates optimization budget to the primitive most relevant to each task, and that same primitive is the best single-component method when run in isolation, but joint optimization still outperforms any single primitive because task performance depends on coupled effects across components.
This pattern is what distinguishes \methodname from single-primitive methods: they can capture the dominant primitive's contribution but not the residual gain from joint search.

\begin{figure}[t]
\centering
\includegraphics[width=\linewidth]{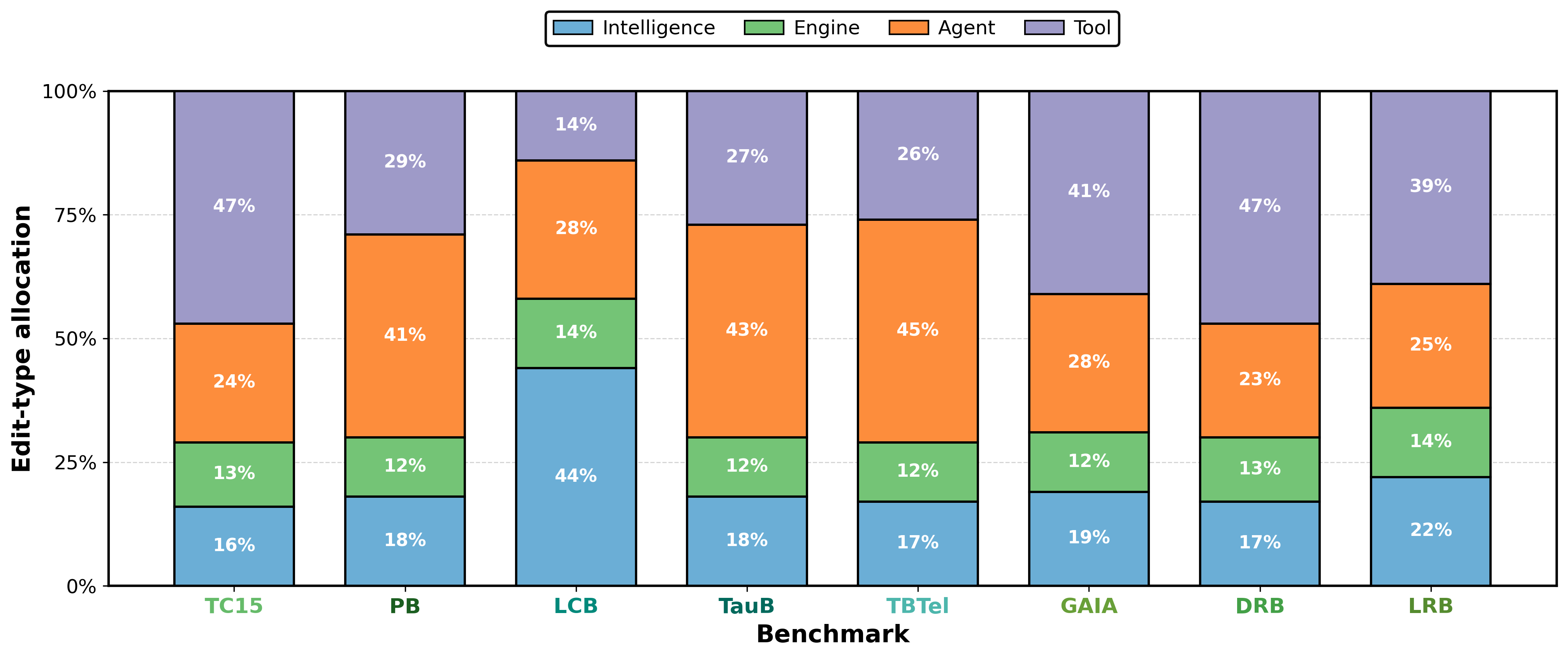}
\caption{\textbf{Edit-type allocation by benchmark.}
Share of accepted edits by primitive across the 8-benchmark suite.
Student: \texttt{Qwen3.5-9B}; teacher: \claudeOpus.
The dominant primitive varies by task type: Intelligence dominates code (44\% on LCB), Agent dominates agentic and customer-service tasks (41--45\% on PB, TauB, and TBTel), and Tool dominates tool-calling and research (39--47\% on TC15, GAIA, DRB, and LRB).
Engine edits appear across benchmarks but primarily affect efficiency rather than standalone accuracy.
Bars are normalized within benchmark.}
\label{fig:edit_type_allocation}
\end{figure}

\begin{figure}[t]
\centering
\includegraphics[width=0.85\linewidth]{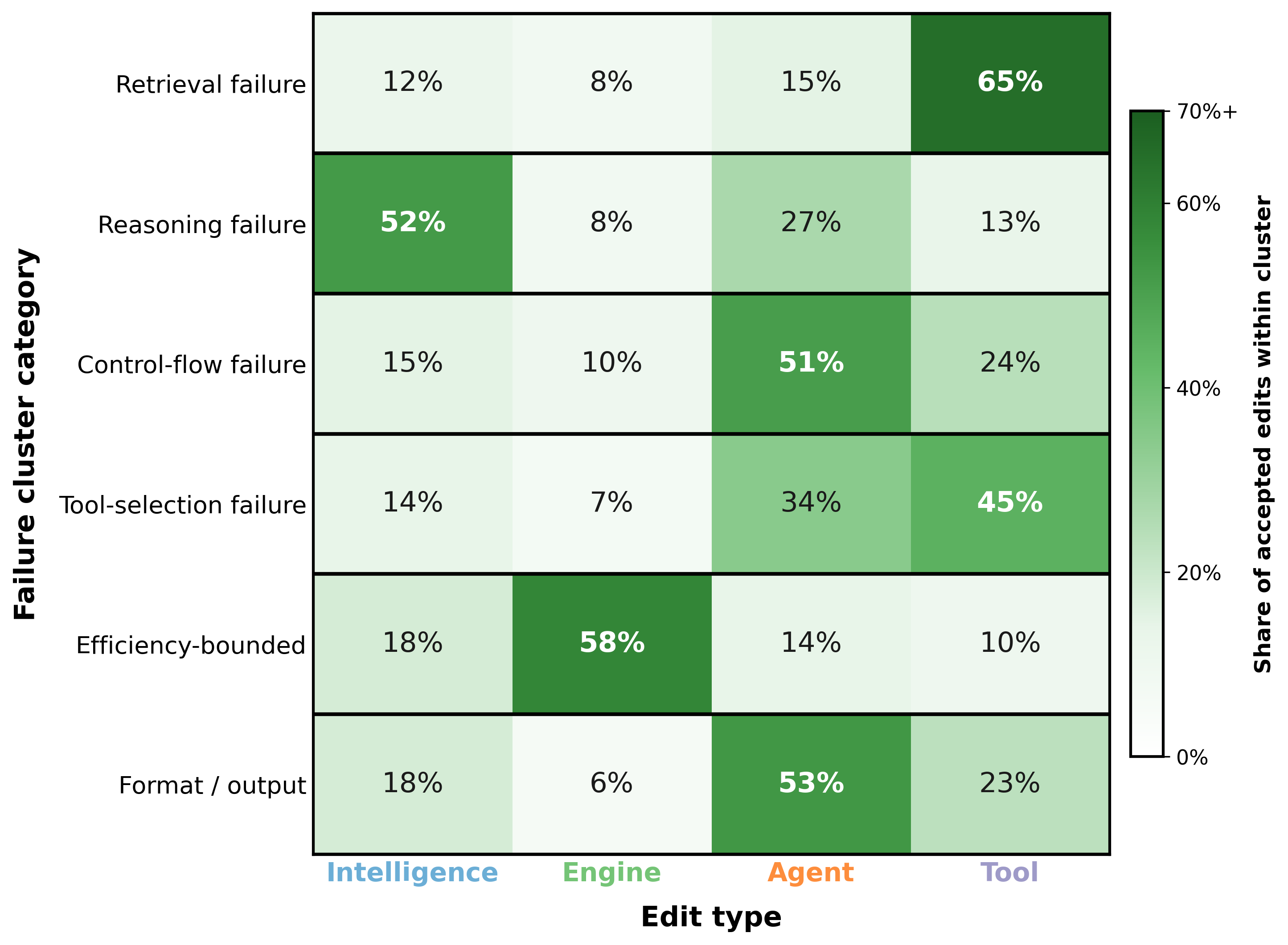}
\caption{\textbf{Edit-type allocation by failure cluster category.}
Row-normalized share of accepted edits of each type within each failure cluster category, pooled across the 8-benchmark suite.
Student: \texttt{Qwen3.5-9B}; teacher: \claudeOpus.
The teacher maps diagnoses to the expected intervention type: retrieval failures receive mostly Tool edits (65\%), reasoning failures mostly Intelligence edits (52\%), control-flow failures mostly Agent edits (51\%), efficiency-bounded failures mostly Engine edits (58\%), and format/output failures mostly Agent edits (53\%).
This pattern supports the claim that the diagnose phase produces semantically meaningful failure clusters rather than arbitrary edit allocation.}
\label{fig:cluster_edit_type}
\end{figure}

\subsection{Search Robustness: Reward Weights, Seed Variance, and Random Restarts}
\label{app:distillation_robustness}

The 13--32~pp search gain reported in Section~\ref{sec:distillation_results} is the headline empirical claim of this paper, and reviewers may reasonably ask whether the gain reflects favorable hyperparameter choices or seed-level variance rather than the \methodname algorithm itself.
This appendix tests those hypotheses.
Together with the editable-set ablation in Appendix~\ref{app:edit_type_ablation}, the studies here cover four independent perturbation axes: \emph{which primitives are editable} (Table~\ref{tab:edit_type_ablation}), \emph{how candidate edits are scored during Intelligence training} (Table~\ref{tab:reward_weight_sensitivity}), \emph{which RNG seed governs edit proposal and acceptance} (Table~\ref{tab:distillation_seed_variance}), and \emph{whether greedy search benefits from random restarts} (Table~\ref{tab:restarts}).
None of the four axes produces variation comparable to the 13--32~pp headline effect.

\paragraph{Reward normalization.}
For each efficiency quantity $X \in \{E,L,C\}$ in Equation~\ref{eq:reward}, we compute $\hat{X}=(X-\mu_X)/\sigma_X$ within the evaluated benchmark before weighting.
The reward therefore trades off dimensionless deviations rather than raw joules, seconds, and dollars.
The reward is used only inside Intelligence edits that trigger training.
The gate score evaluates the edited spec end-to-end.

\paragraph{Reward weight sensitivity.}

\begin{table}[ht]
\centering
\caption{\textbf{Reward weight sensitivity for \methodname.} 
Student: \texttt{Qwen3.5-9B}. Teacher: \texttt{Claude Opus 4.6}. Benchmark: LiveCodeBench. 
Default weights $(\alpha, \beta, \gamma, \delta) = (0.5, 0.1, 0.1, 0.3)$ (Equation~\ref{eq:reward}, bold). 
Each row varies one weight while holding the other three at their defaults; weights are not 
renormalized. Accuracy is mean $\pm$ std dev over 5 runs. The default falls within 2.4 pp of the 
best-performing variant, and all variants land within a 4.6 pp band, indicating the choice of 
defaults does not drive the headline result.}
\label{tab:reward_weight_sensitivity}
\begin{tabular}{lcccccc}
\toprule
Variant & $\alpha$ & $\beta$ & $\gamma$ & $\delta$ & Accuracy (\%) & $\Delta$ vs.\ default \\
\midrule
\multicolumn{7}{l}{\textit{Accuracy weight $\alpha$ (default 0.5)}} \\
Low accuracy weight  & 0.3 & 0.1 & 0.1 & 0.3 & 80.7 $\pm$ 1.6 & $-$2.3 \\
\textbf{Default}     & \textbf{0.5} & \textbf{0.1} & \textbf{0.1} & \textbf{0.3} & \textbf{83.0 $\pm$ 1.3} & \textbf{0.0} \\
High accuracy weight & 0.7 & 0.1 & 0.1 & 0.3 & 84.1 $\pm$ 1.5 & $+$1.1 \\
\midrule
\multicolumn{7}{l}{\textit{Energy penalty $\beta$ (default 0.1)}} \\
No energy penalty    & 0.5 & 0.0 & 0.1 & 0.3 & 82.4 $\pm$ 1.4 & $-$0.6 \\
High energy penalty  & 0.5 & 0.3 & 0.1 & 0.3 & 81.8 $\pm$ 1.7 & $-$1.2 \\
\midrule
\multicolumn{7}{l}{\textit{Latency penalty $\gamma$ (default 0.1)}} \\
No latency penalty   & 0.5 & 0.1 & 0.0 & 0.3 & 83.6 $\pm$ 1.2 & $+$0.6 \\
High latency penalty & 0.5 & 0.1 & 0.3 & 0.3 & 80.9 $\pm$ 1.8 & $-$2.1 \\
\midrule
\multicolumn{7}{l}{\textit{Cost penalty $\delta$ (default 0.3)}} \\
Low cost penalty     & 0.5 & 0.1 & 0.1 & 0.1 & 82.7 $\pm$ 1.3 & $-$0.3 \\
High cost penalty    & 0.5 & 0.1 & 0.1 & 0.5 & 82.2 $\pm$ 1.6 & $-$0.8 \\
\midrule
Accuracy-only ($\alpha$=1, others=0) & 1.0 & 0.0 & 0.0 & 0.0 & 85.4 $\pm$ 1.1 & $+$2.4 \\
\bottomrule
\end{tabular}
\end{table}

Table~\ref{tab:reward_weight_sensitivity} varies each of the four composite-reward weights $(\alpha, \beta, \gamma, \delta)$ in Equation~\ref{eq:reward} around the default $(0.5, 0.1, 0.1, 0.3)$, holding the other three weights fixed.
We evaluate on \texttt{LiveCodeBench} because it is the benchmark where Intelligence edits dominate the accepted-edit distribution (44\% of accepted edits, Table~\ref{tab:edit_type_ablation}), making reward-weight choice most consequential.
All ten weight settings produce final accuracy in a 4.6~pp band (80.7\%--85.4\%), well below the 13--32~pp search gain over the unoptimized baseline (48.5\%, Table~\ref{tab:main_table_results}).
The accuracy weight $\alpha$ has the largest effect (3.4~pp range across its three settings), consistent with $\alpha$ directly weighting the accuracy reward; perturbations to the energy ($\beta$), latency ($\gamma$), and cost ($\delta$) penalties produce changes within the 5-run evaluation noise reported in Appendix~\ref{app:protocol}.
The accuracy-only configuration ($\alpha{=}1$, others $=0$) reaches 85.4\%, 2.4~pp above the default; the composite reward therefore trades 2.4~pp of accuracy in exchange for selecting \ojSpecs that also score well on energy, latency, and cost. This is the same Pareto framing applied at the configuration level in Section~\ref{sec:efficiency}.
Two design implications follow.
First, the headline 13--32~pp gain is not an artifact of the specific weight choice: every reasonable weighting in the table produces gains in the same range.
Second, the default $(0.5, 0.1, 0.1, 0.3)$ is a non-extreme point in this band rather than a tuned optimum, which reduces the risk that the reported numbers reflect implicit hyperparameter search.
Deployments that prioritize accuracy above all else can move toward $\alpha{=}1$ at a roughly 2~pp accuracy gain; deployments that prioritize energy or cost can increase $\beta$ or $\delta$ at a roughly 1~pp accuracy cost.

\paragraph{Search seed variance.}

\begin{table}[ht]
\centering
\caption{\textbf{Search seed variance.} Final accuracy (\%) across 3 search seeds for 
\texttt{Qwen3.5-9B} as student, with each of three teachers, on the three benchmarks from 
Section~\ref{sec:distillation_results}. Each cell shows mean $\pm$ std dev across seeds; each seed itself 
is the mean of 5 evaluation runs (matching the protocol in Appendix~\ref{app:protocol}). 
The $k=5$ stagnation criterion produces stable final \ojSpecs: standard deviation across seeds 
averages 1.4 pp, well below the 13--32 pp search gains reported in 
Section~\ref{sec:distillation_results}.}
\label{tab:distillation_seed_variance}
\begin{tabular}{lccc}
\toprule
Teacher & PinchBench & LiveCodeBench & LiveResearchBench \\
\midrule
\texttt{Claude Opus 4.6} & 99.4 $\pm$ 0.7 & 81.7 $\pm$ 1.8 & 85.3 $\pm$ 2.1 \\
\texttt{GPT 5.4}         & 100.0 $\pm$ 0.0 & 80.9 $\pm$ 1.6 & 89.5 $\pm$ 1.4 \\
\texttt{Gemini 3.1 Pro}  & 99.7 $\pm$ 0.5 & 78.4 $\pm$ 2.3 & 80.6 $\pm$ 1.9 \\
\midrule
Mean across teachers     & 99.7 $\pm$ 0.4 & 80.3 $\pm$ 1.9 & 85.1 $\pm$ 1.8 \\
\bottomrule
\end{tabular}
\end{table}

\begin{table}[t]
\centering
\small
\caption{\textbf{Random restarts ablation.}
Student: \texttt{Qwen3.5-9B}; teacher: \texttt{Claude Opus 4.6}.
\textit{Single} runs \methodname once with stagnation threshold $k$.
\textit{Best-of-3} and \textit{Best-of-5} run independent restarts and report the best final spec by gate score.
Mean over 5 evaluation runs per cell, $\pm$ standard deviation.
The 0.0--2.1~pp gap between Single ($k{=}5$) and Best-of-5 is 1.2~pp on average, comparable to the 1.4~pp seed variance reported in Table~\ref{tab:distillation_seed_variance}.
Thus random restarts do not reveal a large hidden optimization gap under our budget.
This supports the empirical view that useful configurations are often reachable by single LLM-proposed compound edits (Section~\ref{sec:discussion}).}
\label{tab:restarts}
\begin{tabular}{lcccc}
\toprule
Configuration & PinchBench & LiveCodeBench & LiveResearchBench & DeepResearchBench \\
\midrule
Single ($k{=}5$, default) & 100.0 $\pm$ 0.0 & 83.0 $\pm$ 1.8 & 91.0 $\pm$ 1.4 & 92.5 $\pm$ 1.2 \\
Single ($k{=}10$)         & 100.0 $\pm$ 0.0 & 83.9 $\pm$ 1.5 & 91.5 $\pm$ 1.3 & 92.8 $\pm$ 1.1 \\
Best-of-3 ($k{=}5$)       & 100.0 $\pm$ 0.0 & 84.4 $\pm$ 1.2 & 92.1 $\pm$ 1.0 & 93.3 $\pm$ 1.0 \\
Best-of-5 ($k{=}5$)       & 100.0 $\pm$ 0.0 & 85.1 $\pm$ 1.1 & 92.6 $\pm$ 0.9 & 93.6 $\pm$ 0.9 \\
\bottomrule
\end{tabular}
\end{table}

Table~\ref{tab:distillation_seed_variance} reports the variance of final spec accuracy across three independent search seeds for \texttt{Qwen3.5-9B} as student, with each of three teachers, on the three benchmarks from Section~\ref{sec:distillation_results}.
Each seed corresponds to a fresh search run with a different RNG seed governing edit proposal and acceptance; each seed's reported accuracy is itself the mean of 5 evaluation runs (matching the protocol in Appendix~\ref{app:protocol}), so the seed-level standard deviations report variance over and above evaluation noise.
Across the nine (teacher, benchmark) cells, seed-level standard deviation ranges from 0.0--2.3~pp with mean 1.4~pp.
\texttt{PinchBench} shows the lowest variance (0.0--0.7~pp) because all three teachers saturate near 100\%, a ceiling effect we already note in Appendix~\ref{app:limitations}; \texttt{LiveCodeBench} and \texttt{LiveResearchBench} show 1.4--2.3~pp standard deviation, comparable to the 5-run evaluation noise on those benchmarks.
The seed-averaged means align with the headline numbers reported in Section~\ref{sec:distillation_results} and Table~\ref{tab:full_distillation} to within 1.5~pp on every cell, indicating that the headline numbers are typical rather than lucky-seed outliers; for example, Section~\ref{sec:distillation_results} reports \claudeOpus producing an 83\% \texttt{Qwen3.5-9B} student on \texttt{LiveCodeBench}, and Table~\ref{tab:distillation_seed_variance} reports a seed-mean of 81.7\% with standard deviation 1.8~pp on the same configuration.
The highest-variance cell (\texttt{Gemini~3.1~Pro} on \texttt{LiveCodeBench}, std 2.3~pp) remains far below the 13--32~pp search gains reported in Section~\ref{sec:distillation_results}, so seed choice is not a plausible explanation for the headline result on any cell.
The mean seed standard deviation of 1.4~pp is also smaller than the 5.5--16.5~pp gap between full \methodname and the best single-primitive baseline (Figure~\ref{fig:editable_set_ablation}), confirming the contribution from optimizing across primitives to accuracy is real rather than within seed noise.

\paragraph{Combined robustness picture.}
Across the four perturbation axes tested in this paper, the headline 13--32~pp gain survives reasonable variation in editable primitive set (5.5--16.5~pp gap to best solo, Figure~\ref{fig:editable_set_ablation}), reward weights (4.6~pp band across Table~\ref{tab:reward_weight_sensitivity}), search seeds (1.4~pp mean standard deviation, Table~\ref{tab:distillation_seed_variance}), and random restarts (1.2~pp Best-of-5 gain over the default, Table~\ref{tab:restarts}).
None of the four axes produces variation comparable to the headline effect.
The remaining sources of variance not addressed empirically here, especially LLM judge bias, are flagged as limitations in Appendix~\ref{app:limitations}.

\subsection{Proposer Ablation Details}
\label{app:proposer_ablation}

Figure~\ref{fig:editable_set_ablation} compares three proposers at fixed four-primitive move space: a template-random proposer, an evolutionary spec-search proposer over the full spec, and \methodname.
This appendix specifies the baselines.

\paragraph{Shared edit catalog.}
All proposers operate over the same four-primitive edit catalog.
Intelligence templates include model selection, generation-parameter changes, quantization changes, and SFT/LoRA/GRPO training triggers.
Engine templates include backend selection, batch-size changes, KV-cache settings, and runtime-specific optimization flags.
Agent templates include prompt rewrites, few-shot exemplar edits, agent-type changes, verification-policy changes, and turn-limit changes.
Tool templates include tool add/remove decisions, tool-description rewrites, memory-backend changes, and cloud-as-tool routing toggles.
All proposers use the same student, teacher, trace corpus, gate, budget, and evaluation protocol.

\paragraph{Template-random proposer.}
The template-random proposer samples an edit template from the shared catalog, samples its parameters from the allowed range for that template, applies the candidate edit, and uses the same held-out gate as \methodname.
It receives no trace diagnosis and does not condition on failure clusters.
This baseline isolates whether the gains come merely from the edit catalog and gate.

\paragraph{Evolutionary spec-search proposer.}
This baseline keeps the reflective mutation, population, merge, and Pareto-frontier structure of GEPA~\cite{agrawal2026gepareflectivepromptevolution}, but extends the candidate representation from prompt components to the four editable spec primitives.
Each candidate is a serialized spec containing Intelligence, Engine, Agent, and Tool fields.
Mutation proposes a change to one selected primitive using the same trace and gate feedback available to the LLM-guided method.
Merge combines two frontier candidates according to the implementation described in the released code.
The metric-call budget is matched to \methodname.

\paragraph{\methodname.}
\methodname uses the same teacher, trace corpus, edit catalog, gate, and budget, but removes the population and merge machinery.
The teacher may propose compound edits spanning multiple primitives in one step, and the held-out gate accepts the resulting spec only if the target cluster improves without unacceptable regression on other clusters.

\paragraph{Result.}
At fixed four-primitive move space, \methodname improves over evolutionary spec search by 5.5--18.0~pp across PB, LCB, LRB, and DRB, or 10.0~pp on average.
It also improves over the template-random proposer by 7.5--21.0~pp, or 14.0~pp on average.
This isolates proposer and acceptance dynamics from the move-space expansion tested in Figure~\ref{fig:editable_set_ablation}.

\section{Limitations}
\label{app:limitations}

Several limitations qualify the results in this paper.

\paragraph{Benchmark ceiling effects.}
\texttt{PinchBench} saturates at 100\% for both cloud and multiple local models, so the tie on that benchmark reflects ceiling effects rather than true local-cloud parity.
We report it for completeness but do not weight it heavily in our headline claims.

\paragraph{Cloud baseline anomalies.}
Some cloud baselines exhibit unexpectedly low scores on specific benchmarks, which we attribute to incomplete tool integration in our evaluation harness rather than model failure.
Appendix~\ref{app:protocol} documents the diagnostic.
This anomaly does not affect the primary local-vs-cloud comparisons, which are anchored by \claudeOpus and \texttt{GPT~5.4}.

\paragraph{Statistical precision.}
Five independent runs per configuration provide limited precision for sub-5~pp accuracy differences.
Appendix~\ref{app:protocol} reports confidence intervals.
The 3.2~pp average gap between the best local and best cloud model (Section~\ref{sec:portability}) should be interpreted with this precision in mind.

\paragraph{Reward weight sensitivity.}
The composite reward weights $(\alpha, \beta, \gamma, \delta) = (0.5, 0.1, 0.1, 0.3)$ are defaults rather than swept values.
An ablation over reward weights (Table~\ref{tab:reward_weight_sensitivity}) shows accuracy varies by $\pm$2.3~pp across reasonable settings on \texttt{LiveCodeBench}, with the default falling 2.4~pp below an accuracy-only configuration. This is a deliberate trade in exchange for selecting \ojSpecs that are also Pareto-good on energy, latency, and cost (Appendix~\ref{app:distillation_robustness}).

\paragraph{LLM judge bias.}
Scoring uses \texttt{GPT-5-mini} as the LLM judge, which may systematically favor GPT-family outputs.
We have not yet conducted cross-judge validation with Claude or human raters; the headline ranking between the best local model and \claudeOpus (3.2~pp gap, Section~\ref{sec:efficiency}) should be interpreted with this in mind.
We note that Section~\ref{sec:distillation_results} reports \claudeOpus as the strongest teacher on \texttt{LiveCodeBench} despite the GPT-family judge, which is the inverse of the direction judge bias would predict.

\paragraph{Search convergence.}
The $k{=}5$ stopping condition for \methodname may produce local optima.
Variance across 3 search seeds averages 1.4~pp (Table~\ref{tab:distillation_seed_variance}), and Best-of-5 random restarts improve over the default single run by only 1.2~pp on average (Table~\ref{tab:restarts}).
Both effects are well below the 13--32~pp search gains reported in Section~\ref{sec:distillation_results}.
We do not claim a formal convergence guarantee; the proposer-and-space and restart ablations show that greedy search is effective empirically under our matched-budget protocol.

\paragraph{Single-machine evaluation.}
All hardware results are collected on single-machine deployments; we do not evaluate distributed or multi-tenant configurations.
Personal AI deployment is single-user by design, so this matches the use case, but readers interested in shared-deployment scenarios should consult standard inference-serving benchmarks~\cite{mlcommons2024}.

\section{Broader Impacts}
\label{app:broader_impacts}
 
\paragraph{Positive impacts.}
\sysname is designed to shift personal AI from cloud-dependent to on-device execution.
This shift has three direct positive consequences.
First, on-device inference eliminates the transmission of personal data to third-party servers, providing privacy guarantees that cloud-based systems cannot offer by construction (Appendix~\ref{app:privacy}).
Second, local execution reduces per-query energy consumption by roughly 1--20$\times$ and dollar cost by approximately 800$\times$ relative to cloud APIs (Section~\ref{sec:efficiency}), which at the scale of hundreds of millions of daily personal AI users could meaningfully reduce the aggregate energy footprint of AI inference.
Third, by making personal AI functional without a network connection, \sysname extends access to users in low-connectivity environments and reduces dependence on centralized infrastructure whose capacity is increasingly constrained~\cite{bcd2025computecrunch, pilz2025rand}.
 
\paragraph{Privacy and surveillance risk.}
The Tools \& Memory primitive, including Memory, Connectors, and Channels, gives \sysname access to a user's email, messages, calendar, health data, and other personal information.
While this access is necessary for a personal AI to be useful, the same architecture could in principle be repurposed for surveillance or stalking if deployed without the user's informed consent.
Our reference implementation mitigates this risk through mandatory OAuth consent flows for each Connector, audit logs for all cross-Connector queries, and the privacy-by-architecture property that ordinary inference keeps Memory contents and traces on-device (Appendix~\ref{app:privacy}).
When cloud-teacher search is enabled, only eligible scrubbed traces are transmitted during the bounded search phase (Appendix~\ref{app:distillation_privacy}).
We do not support multi-user access to a single \sysname instance's Memory, preventing one user from querying another's personal data.
Deployments in sensitive contexts (healthcare, education, legal) should undergo additional review appropriate to their jurisdiction.
 
\paragraph{Terms of service for teacher APIs.}
\methodname uses frontier cloud models (GPT~5.4, Claude~Opus~4.6, Gemini~3.1~Pro) as teachers that generate labels and propose edits to local \ojSpecs.
The terms of service of OpenAI, Anthropic, and Google variously restrict using model outputs to train competing models.
We note that \methodname primarily updates Agent prompts, Tool configurations, and Engine/runtime settings rather than training a competing foundation model; Intelligence edits that invoke LoRA or GRPO update only a small adapter on top of an independently trained open-weight model.
Nonetheless, users deploying \methodname should verify compliance with their teacher provider's current terms of service before initiating search sessions.
 
\paragraph{Dual-use considerations.}
\sysname is a general-purpose framework for building personal AI agents.
Like any agent framework, it could be used to automate harmful tasks (e.g., generating spam, conducting social engineering, or scraping private data).
The security layer described in Appendix~\ref{app:privacy} provides guardrails (prompt-injection detection, sensitive-data scanning, sandboxed execution), but these are not foolproof.
We release \sysname as open-source software to enable community auditing and improvement of these safeguards.

\end{document}